\pdfoutput=1

\documentclass[11pt]{article}

\usepackage[final]{acl}

\usepackage{times}
\usepackage{latexsym}
\usepackage[T1]{fontenc}
\usepackage[utf8]{inputenc}
\usepackage{microtype}
\usepackage{inconsolata}

\usepackage{graphicx}
\usepackage{subcaption}
\usepackage{booktabs}
\usepackage{multirow}
\usepackage{makecell}
\usepackage{colortbl}
\usepackage{placeins}

\usepackage{amsmath}
\usepackage{amssymb}
\usepackage{bm}
\usepackage{algorithm}
\usepackage{algpseudocode}

\usepackage{enumitem}
\usepackage{pifont}
\usepackage{wasysym}
\usepackage{soul}
\usepackage{color, xcolor}
\usepackage[most]{tcolorbox}
\usepackage{url}
\usepackage{xspace}   %
\definecolor{ao}{rgb}{0.0, 0.5, 0.0}
\definecolor{britishracinggreen}{rgb}{0.0, 0.26, 0.15}
\definecolor{carnelian}{rgb}{0.7, 0.11, 0.11}

\newcommand{\workname}{ProMemAgent\xspace}
\newcommand{\benchmark}{CPMBench\xspace}
\newcommand{\benchmarklite}{CPMBench-Omni\xspace}

\title{\workname: Long-Horizon Proactive Assistance with Context-Aware Prospective Memory from Sensory Perceptions}

\author{
  \textbf{Lilin Xu} \quad
  \textbf{Linyang He} \quad
  \textbf{Kaiyuan Hou} \quad
  \textbf{Yuang Fan} \\
  \textbf{Minghui Zhao} \quad
  \textbf{Xuhai Xu} \quad
  \textbf{Xiaofan Jiang} \\
  Columbia University \\
  \small{\texttt{\{lilin.xu, linyang.he, kh3119, yf2676, mz2866, xx2489\}@columbia.edu}} \\
  \small{\texttt{jiang@ee.columbia.edu}}
}

\begin{document}
\maketitle

\begin{abstract}

Large language model (LLM) agents are moving beyond reactive responses toward proactive assistance that anticipates user needs without explicit requests.
However, existing proactive agents primarily address immediate needs, while many everyday situations involve delayed intentions that can only be acted on hours or days later.
Humans handle such intentions through prospective memory, the ability to remember to act on an intention when the right moment arises.
In this paper, we introduce \workname, the first proactive agent to incorporate prospective memory over everyday sensory contexts for long-horizon proactive assistance.
\workname~organizes past sensory perceptions (e.g., egocentric video and audio) into context-aware memory, generating prospective cues and linking earlier events with subsequent developments for future retrieval.
Given the current context, it retrieves relevant past events and performs memory-augmented proactive reasoning to determine whether assistance is needed.
We further introduce \benchmark, the first benchmark for evaluating long-horizon proactive assistance over sensory contexts, comprising 1,000 instances across eight everyday categories.\footnote{Our dataset and code will be released upon acceptance.}
Extensive experiments show that \workname~achieves the best memory retrieval performance and consistently outperforms all baselines on long-horizon proactive assistance.

\end{abstract}

\section{Introduction}
\label{sec:intro}
Large language model (LLM) agents are moving beyond reactive
responses toward proactive assistance, anticipating user needs and initiating assistance without explicit queries.
Existing proactive agents operate across digital interactions,
such as desktop workflows and mobile interfaces~
\cite{lu2025proactive,lyu-etal-2026-personalalign},
and more recently extend to everyday environments through
sensory contexts captured by wearable devices~
\cite{yang2025contextagent,yang2026proagent}.
However, these agents primarily address needs that can be inferred and acted on at the same moment, using historical information mainly to better infer such needs.
Many assistance needs are instead long-horizon, with their emergence and the opportunity for action separated by hours or days of unrelated daily activities.
For example, a user may run out of oat milk and later pass a grocery store, or need to tell a colleague something and only later run into them at work.

\begin{figure}[t]
    \centering
    \includegraphics[width=\columnwidth]{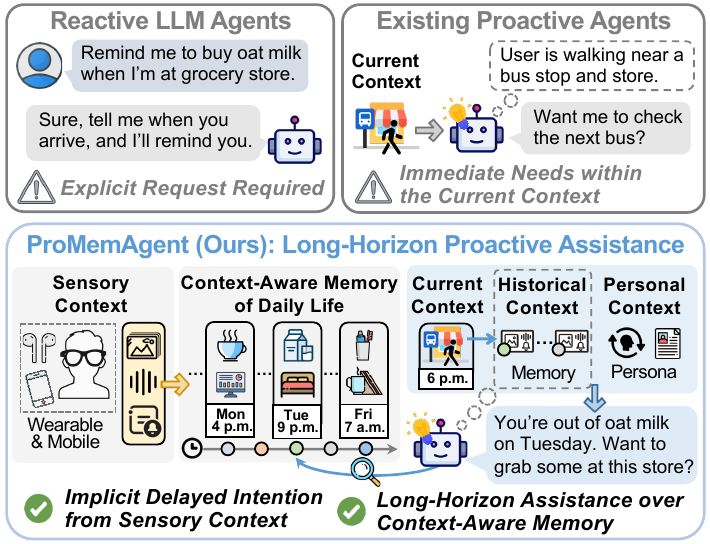}
    \caption{Comparison of prior work (top) and ours (bottom). Reactive agents require explicit requests, and prior proactive agents address immediate needs. 
    \workname~incorporates context-aware prospective memory to connect current and historical contexts and assist with implicit delayed intentions at the right moment.
    }
    \label{fig:teaser}
\end{figure}

This temporal separation closely relates to prospective memory, the human ability to remember to act on a delayed intention at an appropriate future moment~\cite{McDaniel2000S127,einstein2005multiple}.
Recently released agents such as Meta's Muse~\cite{meta2026muse} and OpenAI's Dots~\cite{openai2026dots} follow up on users' goals over time, but still largely rely on users to explicitly articulate their intentions.
Yet many delayed intentions arise naturally in everyday life and are not always stated.
Proactive agents can instead infer such intentions from the user's everyday context, reducing the need for explicit requests.
This motivates \emph{long-horizon proactive assistance}, where agents connect past and current contexts to identify opportunities to act on implicit delayed intentions, as illustrated in Fig.~\ref{fig:teaser}.

Making this connection is challenging.
Recent work shows that LLMs struggle to reliably trigger delayed intentions as context length increases~\cite{zhang2026triggerbench}.
Memory retrieval offers a way to access relevant past information without processing the full history.
However, without an explicit query, the current context serves as the primary retrieval cue and may be only indirectly associated with the earlier context in which a delayed intention arose, limiting similarity-based retrieval.
Recent studies on associative memory move beyond surface similarity~\cite{li-etal-2026-locomo,guo2026tmem}, but still largely rely on explicit queries in conversational settings.
Therefore, a research gap remains in developing proactive agents that can retrieve relevant past events based on the current sensory context and determine when implicit delayed intentions can be acted on.

In this paper, we propose \textbf{\workname}, the first proactive LLM agent to incorporate context-aware \underline{Pro}spective \underline{Mem}ory for long-horizon proactive assistance over sensory contexts.
\workname~first extracts sensory contexts from multimodal perceptions captured by wearable and mobile devices.
It then constructs context-aware memory by generating event-based and time-based prospective cues, and linking earlier events with their subsequent developments.
Given the current context, \workname~retrieves relevant past events through cue-triggered retrieval, and uses a memory-augmented proactive reasoner to decide whether any implicit delayed intention should be acted on.
By connecting past experiences with current opportunities for action, \workname~takes a step toward proactive assistants that serve as external prospective memory in everyday life.

To evaluate long-horizon proactive assistance, we further introduce \benchmark, a \underline{C}ontext-aware \underline{P}rospective \underline{M}emory benchmark containing 1,000 instances across eight everyday categories~(i.e., work, leisure, social, health, food, shopping, transit, and daily life).
We compare \workname with 19 baselines from proactive agents, full-context LLMs, RAG-based methods, and memory systems.
Extensive evaluation shows that \workname~achieves state-of-the-art performance in memory retrieval and long-horizon proactive assistance.
We summarize our contributions as follows.
\begin{itemize}[nosep,leftmargin=*]
\item We introduce the research problem of long-horizon proactive assistance, where implicit delayed intentions arise earlier and become actionable only later, requiring agents to connect the current situation with relevant past context.

\item We propose \workname, the first proactive agent that incorporates context-aware prospective memory over sensory contexts.
It constructs memory with prospective cues and cross-event links for future retrieval, and uses a memory-augmented reasoner to decide whether assistance is needed for any implicit delayed intention.

\item We introduce \benchmark, the first benchmark for long-horizon proactive assistance over sensory contexts.
Extensive experiments show that \workname~achieves the best memory retrieval and outperforms the strongest baseline on proactive assistance by 9.7 points in F1 score.
\end{itemize}

\section{Related Work}
\label{sec:related}
\noindent\textbf{Proactive LLM Agents.}
Proactive agents initiate assistance without explicit user requests.
Existing work anticipates needs from interaction histories in digital environments such as desktop workflows, mobile GUIs, and multi-session conversations~\cite{lu2025proactive,lyu-etal-2026-personalalign,tang2026proagentbench,yang2026fingertip}.
Recent work extends proactive assistance to everyday environments, using sensory contexts from wearable devices to infer user needs~\cite{yang2025contextagent,yang2026proagent}.
However, these studies address immediate needs that arise within the current context, using history mainly to model recurring user patterns that help infer such needs and intervention timing.
More recently, commercial agents such as Meta's Muse~\cite{meta2026muse} and OpenAI's Dots~\cite{openai2026dots} run continuously in the background and follow up on users' goals over time, but largely rely on users to state these goals explicitly.
In contrast, long-horizon proactive assistance over sensory contexts, where needs are inferred from earlier context and addressed when an opportunity arises later, remains underexplored.

\noindent\textbf{Long-Term Memory for LLM Agents.}
Long-term memory enables LLM agents to retain and retrieve information across extended interaction histories.
Existing approaches span lexical and dense retrieval~\cite{robertson2009probabilistic,reimers-2019-sentence}, structured and graph-based representations~\cite{gutierrez2025from,yue-etal-2026-hypermem,rasmussen2025zep}, agentic systems that dynamically construct and update memories~\cite{chhikara2025mem0,xu2026mem,zhong2024memorybank}, and system-level architectures that manage hierarchical memory stores~\cite{packer2023memgpt,kang2025memoryos,li2025memos_long}.
Recent work further evaluates retrieval beyond surface-level semantic similarity.
LoCoMo-Plus~\cite{li-etal-2026-locomo} evaluates memory under cue-trigger semantic disconnect, and T-Mem~\cite{guo2026tmem} introduces a multi-level architecture for both descriptive and associative recall.
However, these approaches remain centered on query-driven recall, leaving open how to trigger relevant memories when the current context is the only cue.

\noindent\textbf{Prospective Memory in LLM Agents.}
Prospective memory, the human ability to retain a delayed intention and act on it when an appropriate condition arises~\cite{McDaniel2000S127,einstein2005multiple}, has recently also been studied in LLMs.
One study~\cite{mittal-2026-forget} examines failures to follow explicit formatting instructions under cognitive load.
For delayed-intention triggering, PM-Bench~\cite{liu2026pm} considers explicitly specified intentions in a text-based simulated week, while TriggerBench~\cite{zhang2026triggerbench} studies implicit prospective constraints introduced earlier in dialogue.
However, existing studies primarily evaluate delayed intentions that are explicitly specified or conveyed through dialogue, leaving prospective memory over rich wearable sensory context underexplored.

\section{Long-Horizon Proactive Assistance}
\label{sec:cpm_task}

\subsection{Task Definition}
\label{sec:task_definition}
Unlike existing proactive agents, which infer immediate needs from the current context, we formalize long-horizon proactive assistance over sensory contexts, where needs arise from prior context and become actionable only at a later opportunity.
At time $t$, an agent $\mathcal{A}$ receives multimodal sensory perceptions $\mathcal{S}_t$, personal context $\mathcal{P}$, and a history $\mathcal{H}_t=(e_1,\ldots,e_{t-1})$ of past events, each capturing the sensory perceptions of an earlier moment.
$\mathcal{S}_t$ comprises egocentric video, audio, and mobile notifications from wearable and mobile devices,
while $\mathcal{P}$ captures relatively stable user information such as habits, preferences, and personal background.
The agent then decides whether any implicit delayed intention should be acted on at the current moment:
$(\mathcal{T},\hat{y},\mathcal{R}) = \mathcal{A}(\mathcal{S}_t,\mathcal{P},\mathcal{H}_t)$.
Here, $\mathcal{T}$ is the reasoning trace and $\hat{y}$ is the proactive decision.
The agent generates a proactive response $\mathcal{R}$ when it decides to act and remains silent otherwise.

\subsection{Task Construction}
\label{sec:task_construction}
To evaluate this task, we introduce \benchmark, the first benchmark for long-horizon proactive assistance over sensory contexts. It consists of daily-life timelines in which implicit delayed intentions are first formed and only later become actionable.

\noindent\textbf{Dataset Design.}
Each instance in \benchmark~consists of personal context $\mathcal{P}$ and a timeline $(e_1,\ldots,e_N)$ of daily-life events, where each event $e_i=(t_i,\mathcal{C}_{t_i})$ pairs a timestamp with sensory context, textual descriptions of the sensory perceptions $\mathcal{S}_{t_i}$ covering egocentric vision, speech, and mobile notifications.
An implicit delayed intention $\mathbf{I}$ is anchored by a \textit{formation moment} $e_f$, where it arises from cues in the sensory context, and a later \textit{decision moment} $e_d$, where the agent decides whether proactive assistance is needed.
The two moments are separated by many intervening daily activities.
A \textit{scenario} specifies $\mathcal{P}$ and one or more intentions formed at different moments, all sharing the final event $e_N$ as their decision moment.
Proactive assistance is needed only if $\mathbf{I}$ has been formed, remains unresolved, and can be acted on at $e_d$.
Each scenario therefore yields four controlled variants, one instance each, as follows:
\begin{itemize}[nosep,leftmargin=*]
\item \texttt{Positive} satisfies all three conditions.
\item \texttt{Negative} offers no opportunity to act at $e_d$.
\item \texttt{State Change} includes an event $e_{sc}$ that resolves $\mathbf{I}$ before $e_d$, so no assistance is needed.
\item \texttt{No Formation} replaces $e_f$ with an ordinary event so no intention arises, but keeps $e_d$.
\end{itemize}
In multi-intention scenarios, variants apply to all intentions jointly.
Together, they test whether an agent can connect the current context to relevant history and determine whether a delayed intention exists, remains unresolved, and is actionable.

\noindent\textbf{Data Construction.}
We construct our dataset in three stages.
First, human annotators write 100 seed scenarios drawn from everyday experiences, each cross-reviewed by another annotator.
We then use LLMs to generate a larger and more diverse set of candidate scenarios, followed by human verification for plausibility and consistency.
Finally, for each scenario, we build the four variant timelines by inserting intention-independent daily-life events before $e_f$ and between $e_f$ and $e_d$.
These events serve as distractors and are shared across the variants, enabling a controlled comparison of the variant-specific events.
This process yields \benchmark, comprising 1,000 instances.

\noindent\textbf{Sensor Data Pairing.}
We further pair a subset of \benchmark with sensor data to form \benchmarklite.
For each event, we retrieve a matching clip from S-EMBER~\cite{wang2026sember}, a large-scale egocentric video dataset, using its visual description, and synthesize audio from its speech text.
When no clip reliably matches or recorded audio is required, we record egocentric video and audio ourselves.
All videos are captured with Meta smart glasses~\cite{meta_rayban}, and the resulting instances are human-verified.
Details on benchmark construction and data pairing are in Appendix~\ref{sec:appendix_construction}.

\begin{figure}[t]
    \centering
    \includegraphics[width=\columnwidth]{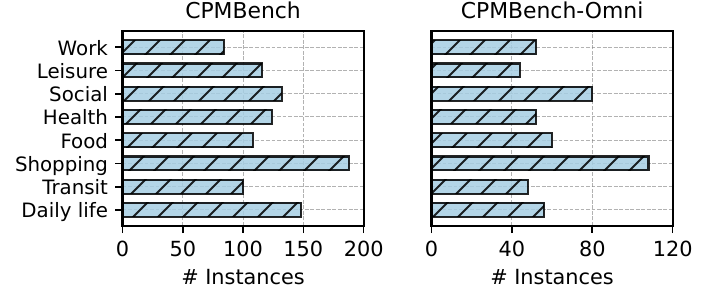}
    \caption{Distribution of timeline instances.}
    \label{fig:scenario_dist}
\end{figure}

\noindent\textbf{Dataset Statistics.}
\benchmark~comprises 1,000 instances derived from 250 scenarios, of which 200 are single-intention and 50 are multi-intention.
As shown in Fig.~\ref{fig:scenario_dist}, the instances involve delayed intentions across eight everyday categories. 
Each instance contains $N=200$ events, with an average of 114.5 events between the formation and decision moments.
\benchmarklite~comprises 500 instances with paired sensor data in total.

\section{\workname~Framework}
\label{sec:framework}
\begin{figure*}[t]
    \centering
    \includegraphics[width=\textwidth]{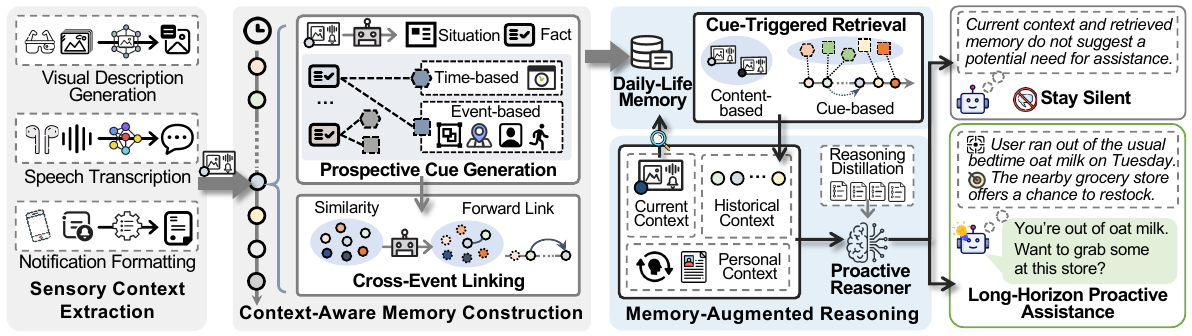}
    \caption{Overview of \workname. It extracts sensory contexts and constructs context-aware memory through prospective cue generation and cross-event linking. During inference, it retrieves past events via cue-triggered retrieval and performs memory-augmented reasoning to provide long-horizon proactive assistance when needed.}
    \label{fig:framework}
\end{figure*}

This section presents \workname, a framework incorporating context-aware prospective memory for long-horizon proactive assistance over sensory contexts.
Fig.~\ref{fig:framework} shows the overview of \workname.
It first extracts sensory contexts from multimodal sensory perceptions and then organizes them into context-aware memory for future retrieval.
During inference, it retrieves relevant past events using the current context and performs memory-augmented proactive reasoning to decide whether proactive assistance is needed.

\subsection{Context-Aware Memory}

\subsubsection{Sensory Context Extraction}
At each moment $t$, \workname~converts the multimodal sensory perceptions $\mathcal{S}_t$ into the sensory context $\mathcal{C}_t$, which together with its timestamp forms an event $e_t=(t,\mathcal{C}_t)$.
A vision-language model (VLM) generates a visual description of the egocentric video, speech is transcribed with speaker attribution to distinguish the user from others, and mobile notifications are formatted as text.
Each moment thus has a unified representation $e_t$ that serves both as the retrieval cue at the current moment and as the entry organized into memory in subsequent moments.

\subsubsection{Memory Construction and Retrieval}
\label{sec:method_memory}
Given each event $e_t$, \workname~employs a memory-construction LLM to organize it into memory for future retrieval.
Without explicit user queries, long-horizon proactive assistance relies on the current sensory context as the only cue for recalling past events.
In daily life, however, the current context and the relevant past event are often semantically disconnected, making similarity-based retrieval insufficient.
Moreover, delayed intentions are usually implicit in sensory contexts and rarely stated explicitly (e.g., ``remind me to buy milk'').
As a result, it is often unclear when an event is observed whether it will later call for assistance, and subsequent events may also change the answer.
Inspired by prospective memory, in which a delayed intention is retrieved when its associated cue occurs~\cite{McDaniel2000S127,einstein2005multiple}, \workname~therefore associates events with prospective cues describing when they may become relevant, links them to their subsequent developments, and retrieves them once the current context triggers these cues.
These three steps are realized by prospective cue generation, cross-event linking, and cue-triggered retrieval, while the proactive assistance decision is left to the reasoner during inference (\S~\ref{sec:reasoning}).

\noindent\textbf{Prospective Cue Generation.}
Motivated by the two primary forms of prospective memory~\cite{einstein1990normal}, \workname~generates event-based and time-based prospective cues.
For each stored event $e_i$, it summarizes the observed situation as $q_i$ and extracts a set of factual observations $\mathcal{F}_i$ from its sensory context.
For each fact $f\in\mathcal{F}_i$, it constructs a set of event-based cues $\mathcal{K}_f$, each describing a future situation in which the fact may become useful, and a set of time-based cues $\mathcal{W}_f$, each specifying a future time point or window.
Observed situations and event-based cues share a structured form covering the user's activity, location, surrounding context, and people involved, following cognitive models of event representation~\cite{radvansky2011event}.
This shared form allows a cue to be compared directly with the situation of a later event.
The cues are used for memory retrieval, while the original event $e_i$ remains the evidence for proactive reasoning.

\noindent\textbf{Cross-Event Linking.}
In human prospective memory, a finished intention is deactivated and no longer triggered by its cue~\cite{scullin2009evidence}.
\workname~thus tracks whether earlier observations have been resolved or changed, connecting them with their subsequent developments as new events arrive.
For each new event $e_j$, every fact $f$ extracted from a preceding event $e_i$ is scored by its relevance to $e_j$, taken as the highest of $\mathrm{sim}(\mathcal{C}_j,f)$, $\mathrm{sim}(\mathcal{C}_j,\mathcal{C}_i)$, and $\max_{\kappa\in\mathcal{K}_f}\mathrm{sim}(q_j,\kappa)$, where $\mathrm{sim}$ denotes cosine similarity between text embeddings.
\workname~takes the top-$m$ facts as candidates and uses the memory-construction LLM to determine whether $e_j$ updates each of them.
A cross-event link is created only when the relation is supported by evidence explicitly present in $e_j$.
Links are directed forward in time and added incrementally.
The memory entry of each event $e_i$ is thus
\begin{equation}
\left(e_i,\; q_i,\; \left\{(f,\mathcal{K}_f,\mathcal{W}_f,\mathcal{L}_f)\right\}_{f\in\mathcal{F}_i}\right),
\end{equation}
where $\mathcal{L}_f$ is the set of later events linked from $f$.

\noindent\textbf{Cue-Triggered Retrieval.}
In prospective memory, an intention is recalled when its cue is encountered, without an explicit request~\cite{McDaniel2000S127}.
During inference, \workname~uses the current event $e_t$ to retrieve $k$ past events through complementary content retrieval and cue-triggered retrieval.
Content retrieval directly compares sensory contexts and selects the $k^{C}$ past events with the highest $\mathrm{sim}(\mathcal{C}_t,\mathcal{C}_i)$, denoted $\mathcal{M}^{C}$.
Cue-triggered retrieval instead reaches past events through prospective cues.
Event-based cues act as contextual triggers, so each past event $e_i$ is scored by $\max_{f\in\mathcal{F}_i}\max_{\kappa\in\mathcal{K}_f}\mathrm{sim}(\mathcal{C}_t,\kappa)$.
Time-based cues act as temporal triggers, and an event is time-triggered if $t$ satisfies any time-based cue of its facts.
Events with and without a time trigger are ranked separately and interleaved into one candidate list $\mathcal{Q}$ to balance temporal and contextual relevance.
Each candidate is then expanded with the latest event linked from its facts, when available.
\workname~keeps the top $k^{CT}$ events in $\mathcal{Q}$ that content retrieval has not already selected, denoted $\mathcal{M}^{CT}$.
Together, the $k=k^{C}+k^{CT}$ events in $\mathcal{M}^{C}$ and $\mathcal{M}^{CT}$ are ordered chronologically to form $\mathcal{M}_r$ for proactive reasoning.

\subsection{Memory-Augmented Proactive Reasoning}
\label{sec:reasoning}

Existing proactive agents infer immediate needs from the current context~\cite{lu2025proactive,yang2025contextagent}.
Long-horizon assistance, however, requires interpreting the current context together with relevant past events.
\workname~thus augments proactive reasoning with retrieved memory.

\noindent\textbf{Memory-Augmented Reasoner.}
Given the current sensory context $\mathcal{C}_t$, personal context $\mathcal{P}$, and the retrieved events $\mathcal{M}_r$ as historical context, \workname~develops a proactive reasoner $\mathcal{A}_R$ that produces the reasoning trace, proactive decision, and proactive response, $(\mathcal{T},\hat{y},\mathcal{R})=\mathcal{A}_R(\mathcal{C}_t,\mathcal{P},\mathcal{M}_r)$.
The personal context $\mathcal{P}$ is taken from \benchmark, while in practice it can be extracted from longitudinal sensor streams over months~\cite{yang2026sensorpersona}.
The reasoner decides whether proactive assistance is needed by reasoning about whether an implicit delayed intention has been formed, remains unresolved, and can be acted on at the current moment.

\noindent\textbf{Reasoning Distillation.}
We construct each training context by combining key events with distractor events, so that the reasoner learns to account for historical context in its proactive reasoning.
Specifically, we consider three key-event configurations, $\mathcal{E}\in\{\{e_f\},\{e_f,e_{sc}\},\varnothing\}$, representing cases with formation evidence only, with formation followed by a subsequent development, and without formation evidence.
For each context size $K$, we construct the historical context $\mathcal{H}_K=\mathcal{E}\cup\mathcal{D}_K$ with $|\mathcal{H}_K|=K$, where $\mathcal{D}_K$ contains distractor events randomly sampled from earlier in the timeline.
Varying $\mathcal{E}$ and $K$ diversifies both the reasoning conditions and the amount of historical context.
To obtain reasoning supervision, a frontier LLM takes $(\mathcal{C}_t,\mathcal{P},\mathcal{H}_K)$ and generates an evidence-grounded reasoning trace explaining how they support the target proactive decision and response.
We fine-tune $\mathcal{A}_R$ with LoRA~\cite{hu2022lora} to generate the distilled trace followed by the proactive decision and response.
During inference, $\mathcal{H}_K$ is replaced by the retrieved events $\mathcal{M}_r$, enabling the reasoner to apply the learned reasoning to retrieved memory.

\section{Experiments}
\label{sec:eval}
\subsection{Experimental Setup}
\label{sec:implement_details}

\noindent\textbf{Baselines.}
We compare \workname~against four families of baselines.
\textbf{\textit{Proactive Agents}} include Proactive Agent~\cite{lu2025proactive} and ContextAgent~\cite{yang2025contextagent}, which infer assistance needs without explicitly retrieving relevant past events.
\textbf{\textit{Full-Context}} baselines use open-source and proprietary LLMs~\cite{yang2025qwen3,yang2025qwen25,meta2024llama33,openai_models} to reason directly over the full historical context.
\textbf{\textit{RAG-Based}} baselines retrieve past events using BM25~\cite{robertson2009probabilistic}, all-MiniLM-L6-v2~\cite{reimers-2019-sentence}, or text-embedding-3-small~\cite{openai2024embeddings}.
\textbf{\textit{Memory Systems}} include Mem0~\cite{chhikara2025mem0}, A-Mem~\cite{xu2026mem}, MemOS~\cite{li2025memos_long}, and T-Mem~\cite{guo2026tmem}, which construct and retrieve long-term memories before reasoning.

\noindent\textbf{Metrics.}
We evaluate long-horizon proactive assistance along three dimensions.
\textbf{\textit{Memory Retrieval}.}
Delayed intention recall \textbf{(DIR)} is the fraction of delayed intentions in \texttt{Positive} instances whose formation moment $e_f$ is retrieved, and state change recall \textbf{(SCR)} is the fraction of \texttt{State Change} instances in which both $e_f$ and the state-change event $e_{sc}$ are retrieved.
\textbf{\textit{Proactive Decision}.}
Following prior work~\cite{yang2025contextagent,lu2025proactive}, we use false detection rate \textbf{(FD)} and proactive prediction accuracy \textbf{(Acc-P)} to evaluate whether the agent correctly decides to assist or remain silent.
\textbf{\textit{End-to-End Assistance}.}
We further evaluate response content using an LLM judge~\cite{zheng2023judging,li2024llmsasjudges}.
Since an instance may contain multiple delayed intentions, we use set-level precision \textbf{(Set-P)}, recall \textbf{(Set-R)}, and F1 \textbf{(Set-F1)}.
Proactive assistance accuracy \textbf{(Acc-PA)} extends Acc-P by also requiring the assistance to be correct.
Metric details are in Appendix~\ref{sec:appendix_metrics}.

\noindent\textbf{Implementation Details.}
We split \benchmark by scenario into 60\% training and 40\% testing, so all variants of a scenario share a split.
For sensory context extraction, \workname~uses Qwen3-VL-32B and WhisperX~\cite{bain2022whisperx}.
Unless otherwise specified, \workname~and proactive-agent baselines use Qwen3-8B, fine-tuned with LoRA under the same training settings.
Reasoning traces for distillation are generated by Claude-Opus-4.6.
By default, applicable methods use GPT-4o-mini for LLM-based memory construction and text-embedding-3-small for embeddings.
We set $k=10$ for top-$k$ retrieval methods except T-Mem, for which we evaluate its default setting and a variant that retrieves the top-$10$ scenes using only horizon triggers.
\workname~uses $m=8$ for cross-event linking, $k^C=2$ for content retrieval, and $k^{CT}=8$ for cue-triggered retrieval.
RAG-based and memory-system baselines use GPT-5.4 for proactive reasoning.
We use GPT-5-mini as the LLM judge, which shows high agreement with human annotators (Appendix~\ref{sec:appendix_judge}).
Implementation and prompt details are in Appendices~\ref{sec:appendix_implementation} and~\ref{sec:appendix_prompts}.

\begin{table*}[t]
\centering
\small
\setlength{\tabcolsep}{4pt}
\resizebox{0.95\textwidth}{!}{%
\begin{tabular}{p{2.0cm}lcccccccc}
\toprule
& \textbf{Method}
& \multicolumn{2}{c}{\textbf{Memory Retrieval}}
& \multicolumn{2}{c}{\textbf{Proactive Decision}}
& \multicolumn{4}{c}{\textbf{End-to-End Assistance}} \\
\cmidrule(lr){3-4} \cmidrule(lr){5-6} \cmidrule(lr){7-10}
& & DIR$\uparrow$ & SCR$\uparrow$ & FD$\downarrow$ & Acc-P$\uparrow$ & Acc-PA$\uparrow$ & Set-P$\uparrow$ & Set-R$\uparrow$ & Set-F1$\uparrow$ \\
\midrule
\multirow{2}{=}{\centering Proactive Agents} & Proactive Agent~(8B)  & - & - & 7.0 & 77.8 & 74.7 & 63.8 & 30.1 & 40.9 \\
& ContextAgent~(8B)     & - & - & 43.3 & 59.8 & 46.8 & 17.7 & 22.8 & 19.9 \\

\midrule
\multirow{11}{=}{\centering Full-Context\\LLMs\\(Open- and\\\mbox{Closed-Source})} & Qwen2.5-32B           & - & - & 15.7 & 73.2 & 68.3 & 43.4 & 29.3 & 35.0 \\
& Qwen3-4B              & - & - & 19.0 & 69.8 & 63.8 & 32.1 & 22.0 & 26.1 \\
& Qwen3-8B              & - & - & 27.0 & 69.8 & 63.4 & 37.7 & 39.8 & 38.7 \\
& Qwen3-14B             & - & - & 16.0 & 74.8 & 70.2 & 48.4 & 36.6 & 41.7 \\
& Qwen3-32B             & - & - & 41.3 & 64.8 & 60.0 & 38.6 & 63.4 & 48.0 \\
& Llama-3.3-70B         & - & - & 16.3 & 72.8 & 68.6 & 44.3 & 31.7 & 37.0 \\

\cmidrule(l){2-10}
& GPT-4o                & - & - & 13.0 & 82.0 & 77.3 & 62.9 & 53.7 & 57.9 \\
& GPT-4.1               & - & - & 5.3 & \underline{86.0} & \underline{81.3} & \underline{78.9} & 48.8 & 60.3 \\
& GPT-5.4-mini          & - & - & 5.3 & 77.5 & 73.3 & 61.9 & 21.1 & 31.5 \\
& GPT-5.4               & - & - & 24.0 & 79.8 & 76.4 & 56.9 & \textbf{77.2} & \underline{65.5} \\

\midrule
\multirow{3}{=}{\centering RAG-based Methods (GPT-5.4)} & all-MiniLM-L6-v2      & 14.6 & 10.0 & \textbf{3.0} & 77.2 & 73.0 & 66.7 & 14.6 & 24.0 \\
& BM25                  & 19.5 & 13.0 & 5.0 & 77.0 & 72.8 & 60.5 & 18.7 & 28.6 \\
& text-embedding-3-small & 21.1 & 13.0 & 6.3 & 76.8 & 72.1 & 55.8 & 19.5 & 28.9 \\

\midrule
\multirow{5}{=}{\centering Memory Systems (GPT-5.4)} & Mem0                  & - & - & 6.3 & 74.5 & 70.2 & 45.7 & 13.0 & 20.3 \\
& A-Mem                 & 32.5 & \underline{25.0} & 7.3 & 79.5 & 74.9 & 63.9 & 31.7 & 42.4 \\
& MemOS                 & 30.9 & 6.0 & 12.0 & 74.8 & 70.2 & 47.8 & 26.8 & 34.4 \\
& T-Mem                 & 28.5 & 19.0 & 9.0 & 74.5 & 70.2 & 47.1 & 19.5 & 27.6 \\
& T-Mem (Horizon-based) & \underline{49.6} & \underline{25.0} & 15.0 & 75.0 & 70.9 & 50.0 & 36.6 & 42.3 \\

\midrule
\multirow{1}{=}{\centering Ours} & \workname~(8B)        & \textbf{81.3} & \textbf{72.0} & \underline{4.3} & \textbf{90.0} & \textbf{87.2} & \textbf{86.3} & \underline{66.7} & \textbf{75.2} \\

\bottomrule
\end{tabular}%
}
\caption{Main results on \benchmark. All metrics are in percentages (\%). The best and second-best results are in \textbf{bold} and \underline{underlined}, respectively. ``-'' denotes inapplicable retrieval metrics.}
\label{tab:main_cpmbench_compact_gpt5mini}
\end{table*}

\noindent\textbf{Overall performance on \benchmark.}
Table~\ref{tab:main_cpmbench_compact_gpt5mini} shows that \workname, with an 8B reasoner, achieves the best overall performance across memory retrieval, proactive decision making, and end-to-end assistance.
It improves Set-F1 by 9.7 points over the strongest baseline, full-context GPT-5.4, indicating its effectiveness for long-horizon proactive assistance.
Proactive agents designed for immediate needs miss most delayed intentions.
Access to the full history is not sufficient either.
With relevant events buried among many unrelated ones, full-context LLMs either overlook many intentions or, like GPT-5.4, achieve higher Set-R at the cost of a high false detection rate.
Similarity-based RAG methods and memory systems rarely reach relevant past events, which are often semantically disconnected from the current context.
Retrieving with T-Mem's horizon triggers alone improves DIR, indicating the value of anticipating future use during memory construction.
\workname~achieves the highest DIR and SCR, highlighting the benefit of prospective cues and cross-event links for retrieving relevant past events from sensory contexts.

\begin{table}[t]
\centering
\small
\setlength{\tabcolsep}{3pt}
\resizebox{\columnwidth}{!}{%
\begin{tabular}{lccccc}
\toprule
\textbf{Method} & DIR$\uparrow$ & SCR$\uparrow$ & Set-P$\uparrow$ & Set-R$\uparrow$ & Set-F1$\uparrow$ \\
\midrule
Proactive Agent~(8B)            & - & - & 63.6 & 21.5 & 32.2 \\
ContextAgent~(8B)               & - & - & 11.0 & 12.3 & 11.6 \\
\midrule
Qwen3-32B        & - & - & 35.2 & 49.2 & 41.0 \\
GPT-5.4          & - & - & 50.0 & 61.5 & 55.2 \\
\midrule
text-embedding-3-small    & 20.0 & 10.7 & 55.6 & 15.4 & 24.1 \\
T-Mem (Horizon-based) & 26.2 & 14.3 & 44.8 & 20.0 & 27.7 \\
\midrule
\workname~(8B)                  & \textbf{83.1} & \textbf{66.1} & \textbf{74.5} & \textbf{58.5} & \textbf{65.5} \\
\bottomrule
\end{tabular}%
}
\caption{Main results on \benchmarklite.}
\label{tab:main_cpmbench_lite_compact}
\end{table}

\begin{figure}[t]
    \centering
    \includegraphics[width=\columnwidth]{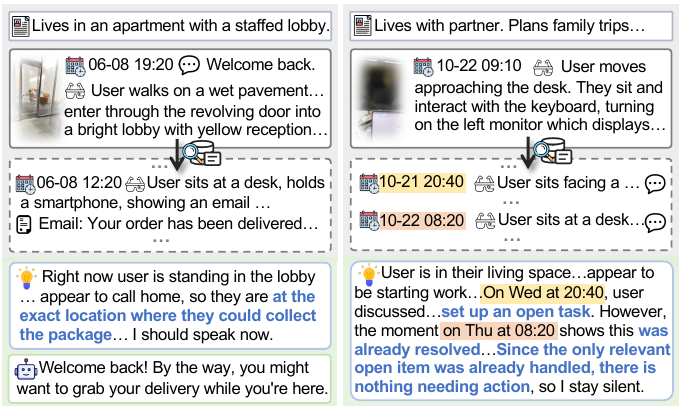}
    \caption{ 
    Qualitative results of \workname in proactive and non-proactive cases.
    }
    \label{fig:result_example}
\end{figure}

\noindent\textbf{Overall performance on \benchmarklite.}
As shown in Tab.~\ref{tab:main_cpmbench_lite_compact} (complete results in Appendix~\ref{sec:appendix_results}), all methods show lower end-to-end performance on \benchmarklite~than on \benchmark, reflecting the additional challenge of imperfect sensory context extraction.
\workname~still maintains its advantage over all baselines and outperforms full-context GPT-5.4 by 10.3 points in Set-F1.
Its high DIR and SCR show that it effectively retrieves both intention-formation events and their subsequent developments from noisy sensory contexts.
Fig.~\ref{fig:result_example} illustrates two cases in which \workname~assists when a delayed intention becomes actionable and remains silent when it has been resolved.

\subsection{Ablation Study and Discussion}
\begin{table}[t]
\centering
\small
\setlength{\tabcolsep}{3pt}
\resizebox{\columnwidth}{!}{%
\begin{tabular}{lcccccc}
\toprule
\textbf{Method}
& \multicolumn{5}{c}{\textbf{\benchmark}}
& \textbf{Direct QA} \\
\cmidrule(lr){2-6} \cmidrule(lr){7-7}
& DIR$\uparrow$ & SCR$\uparrow$ & FD$\downarrow$ & Set-R$\uparrow$ & Set-F1$\uparrow$ & Recall@10$\uparrow$ \\
\midrule
\workname            & 81.3 & 72.0 & 4.3 & 66.7 & 75.2 & 98.1 \\
\midrule
w/o $\mathcal{L}_f$       & 81.3 & 38.0 & 14.7 & 66.7 & 65.9 & 98.1 \\
w/o $\mathcal{K}_f$       & 47.2 & 42.0 & 2.7 & 32.5 & 46.8 & 99.5 \\
w/o $\mathcal{W}_f$       & 66.7 & 61.0 & 3.0 & 54.5 & 67.3 & 98.1 \\
w/o $\mathcal{M}^{CT}$    & 26.0 & 15.0 & 4.0 & 21.1 & 32.3 & 99.5 \\
w/o $\mathcal{M}^{C}$     & 80.5 & 72.0 & 4.3 & 65.0 & 74.1 & 70.0 \\
\bottomrule
\end{tabular}%
}
\caption{Memory components ablation on \benchmark.
}
\label{tab:ablation_memory}
\end{table}

\noindent\textbf{Memory Components.}
Tab.~\ref{tab:ablation_memory} shows that all memory components contribute to \workname's performance.
Removing cue-triggered retrieval ($\mathcal{M}^{CT}$) degrades both retrieval and assistance performance, indicating that relevant past events are often semantically disconnected from current context.
Event-based ($\mathcal{K}_f$) and time-based ($\mathcal{W}_f$) cues provide complementary contextual and temporal triggers.
Cross-event linking ($\mathcal{L}_f$) helps retrieve subsequent developments and avoid acting on resolved intentions.
Content retrieval ($\mathcal{M}^{C}$) complements cue-triggered retrieval by preserving similarity-based access to memory.
On a direct QA set built from \benchmark, where queries ask about information explicitly observed in past events, removing it reduces Recall@10 from 98.1 to 70.0.

\begin{figure}[t]
    \centering
    \begin{subfigure}{0.49\columnwidth}
        \centering
        \includegraphics[width=\textwidth]{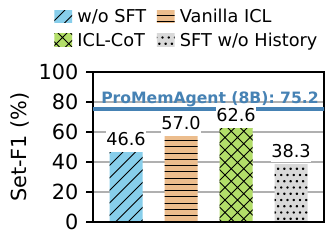}
        \caption{Reasoning components.}
        \label{fig:reasoner_setf1_components}
    \end{subfigure}
    \hfill
    \begin{subfigure}{0.49\columnwidth}
        \centering
        \includegraphics[width=\textwidth]{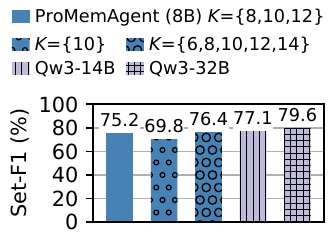}
        \caption{Backbone and $K$.}
        \label{fig:reasoner_setf1_backbone}
    \end{subfigure}
    \caption{Impact of reasoner variants on \benchmark. 
     $K$ denotes training context sizes and ``Qw'' denotes Qwen.
    }
    \label{fig:reasoner_setf1}
\end{figure}

\noindent\textbf{Reasoning Components.}
Fig.~\ref{fig:reasoner_setf1_components} compares reasoner variants given the same retrieved memory.
Even with relevant past events, the base Qwen3-8B reasoner (w/o SFT) struggles with implicit delayed intentions.
In-context learning with 10 demonstrations (Vanilla ICL) improves it, and adding the reasoning traces used for distillation (ICL-CoT) improves it further, showing the value of explicit reasoning.
Fine-tuning on the current context alone (SFT w/o History), as in ContextAgent, performs poorly, indicating the need to train the reasoner with historical context.
On three randomly selected categories excluded from fine-tuning, \workname~still achieves 74.0 Set-F1 and outperforms all baselines (detailed results in Appendix~\ref{sec:appendix_results}), showing the generalizability of its proactive reasoning.

\noindent\textbf{Sensory and Personal Context.}
On \benchmarklite, removing vision or audio at inference lowers Set-F1 by 20.0 and 4.9 points, respectively, showing that both modalities are important.
Without personal context, Set-F1 drops by 5.1 points, mainly in recall, suggesting that personal context helps recognize delayed intentions tied to the user's habits and background.
Replacing the 32B VLM with Qwen3-VL-8B lowers Set-F1 by 6.2 points, suggesting that a stronger VLM for sensory context extraction benefits long-horizon proactive assistance.

\subsection{Sensitivity Analysis}

\noindent\textbf{Memory Construction.}
Memory construction is robust to both the construction LLM and the number of candidate facts $m$ for cross-event linking (Tab.~\ref{tab:analysis_linkm} in Appendix~\ref{sec:appendix_results}).
Replacing GPT-4o-mini with GPT-4.1-mini or varying $m$ between 4 and 16 changes Set-F1 by less than 3 points.

\begin{figure}[t]
    \centering
    \includegraphics[width=\columnwidth]{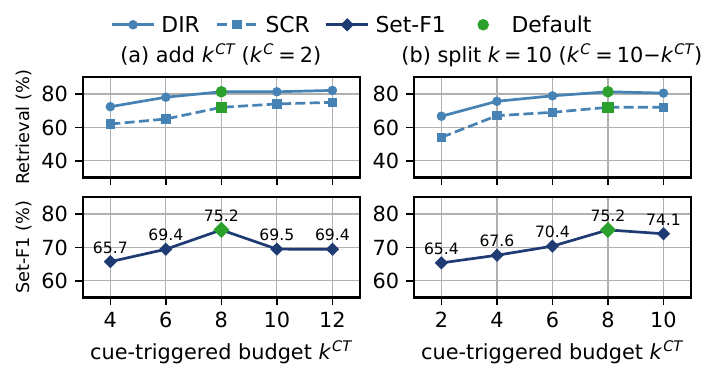}
    \caption{
    Sensitivity to the content and cue-triggered retrieval budgets ($k^{C}$, $k^{CT}$) on \benchmark. 
    }
    \label{fig:retrieval_kct}
\end{figure}

\noindent\textbf{Retrieval Configuration.}
Fig.~\ref{fig:retrieval_kct} varies the cue-triggered budget $k^{CT}$, either with the content budget fixed at $k^{C}=2$ or with the total budget fixed at $k=10$.
In both settings, increasing $k^{CT}$ first improves both retrieval and Set-F1, indicating the need for sufficient cue-triggered retrieval.
Further increasing $k^{CT}$ provides little additional retrieval benefit and can reduce Set-F1, suggesting that extra retrieved events may distract the reasoner.

\noindent\textbf{Reasoner Configuration.}
Fig.~\ref{fig:reasoner_setf1_backbone} varies the reasoner backbone and training context sizes $K$.
Scaling the backbone from Qwen3-8B to Qwen3-32B consistently improves Set-F1, suggesting that stronger reasoners better exploit the same retrieved memory.
Training with a single context size underperforms training with multiple sizes, even when it matches the inference budget $k$, showing the benefit of varying the amount of historical context.

\section{Conclusion}
\label{sec:conclusion}
This paper introduces \workname, the first proactive LLM agent to incorporate context-aware prospective memory for long-horizon proactive assistance.
\workname~leverages sensory contexts for effective memory construction and performs memory-augmented proactive reasoning for implicit delayed intentions.
We also introduce \benchmark, the first benchmark for this task.
Our work extends proactive assistance from immediate needs to delayed intentions, helping users act on them when the opportunity arises in everyday life.

\section*{Limitations}
\label{sec:limitations}
This work evaluates long-horizon proactive assistance over sequences of discrete sensory contexts.
While \benchmarklite~pairs events with real egocentric video and audio, these clips are collected per event, and our work does not model the full pipeline from continuous wearable sensing, such as segmenting continuous streams into events.
Evaluation on continuous in-the-wild wearable streams is left to future work.
Although \benchmark~includes eight everyday categories in English, it can be diversified across categories and languages to further test the generality of long-horizon proactive assistance.
\workname~converts raw video and audio into textual sensory contexts before memory construction and reasoning, which may lose some fine-grained multimodal information.
Preserving and directly reasoning over multimodal representations throughout the memory and reasoning pipeline remains future work.
Memory construction and reasoning distillation further rely on LLMs that can reliably follow instructions, and weaker LLMs may affect the quality of intermediate representations.

\section*{Ethical Considerations}
\label{sec:ethics}
All instances in \benchmark, whether derived from human-written seeds or LLM-generated scenarios, contain no identifiable information about real individuals and include only synthetic personas and names.
The sensory data in \benchmarklite are drawn partly from S-EMBER~\cite{wang2026sember}, a publicly released dataset, and partly from our own data collection.
The S-EMBER videos we use have faces blurred and audio removed, and we use them in accordance with the dataset's license for non-commercial research.
Our data collection was approved by the Institutional Review Board (IRB) of our institution, with informed consent obtained from all participants.
Faces and other sensitive regions are blurred to protect privacy, and speech in the recordings followed scripted scenarios with synthetic names and contains no personal information about participants.
While \workname is evaluated in a controlled setting, a real-world deployment would store long-term sensory contexts from wearable and mobile devices and personal information about the user, potentially including information about bystanders.
Such deployment would require appropriate privacy safeguards, but questions regarding how such memories should be stored, retained, or governed in deployed systems are beyond the scope of this work.

\section*{Acknowledgments}
This research was partially supported by the National Science Foundation under Grant Number CNS-1943396. The views and conclusions contained here are those of the authors and should not be interpreted as necessarily representing the official policies or endorsements, either expressed or implied, of Columbia University, NSF, or the U.S. Government or any of its agencies.

\bibliography{ref}

@misc{azure_tts,
  author       = {{Microsoft Azure}},
  title        = {Text to Speech Overview, Azure Speech Service},
  year         = {2026},
  howpublished = {\url{https://learn.microsoft.com/en-us/azure/ai-services/speech-service/text-to-speech}},
  note         = {Accessed: 2026-10-02}
}

@article{wang2026sember,
  title = {S-EMBER: A Large-Scale Benchmark for Streaming Egocentric Memory Retrieval},
  author = {Wang, Xiaodong and Zhao, Xuanyi and Rodriguez, Pedro and Sachan, Devendra Singh and Oguz, Barlas and Moon, Seungwhan and Li, Shang-Wen and Ghosh, Gargi and Dong, Xin and Yih, Wen-Tau},
  journal = {arXiv preprint arXiv:2607.02689},
  year = {2026},
  url = {https://arxiv.org/abs/2607.02689}
}

@inproceedings{yang2025contextagent,
  title={ContextAgent: Context-Aware Proactive LLM Agents with Open-World Sensory Perceptions},
  author={Yang, Bufang and Xu, Lilin and Zeng, Liekang and Liu, Kaiwei and Jiang, Siyang and Lu, Wenrui and Chen, Hongkai and Jiang, Xiaofan and Xing, Guoliang and Yan, Zhenyu},
  booktitle={Advances in Neural Information Processing Systems},
  volume={38},
  pages={167509--167543},
  year={2025}
}

@inproceedings{lu2025proactive,
  author = {Lu, Yaxi and Yang, Shenzhi and Qian, Cheng and Chen, Guirong and Luo, Qinyu and Wu, Yesai and Wang, Huadong and Cong, Xin and Zhang, Zhong and Lin, Yankai and Liu, Weiwen and Wang, Yasheng and Liu, Zhiyuan and Liu, Fangming and Sun, Maosong},
  booktitle = {International Conference on Learning Representations},
  editor = {Y. Yue and A. Garg and N. Peng and F. Sha and R. Yu},
  pages = {47431--47457},
  title = {Proactive Agent: Shifting LLM Agents from Reactive Responses to Active Assistance},
  url = {https://proceedings.iclr.cc/paper_files/paper/2025/file/75c37811e830bf029584b1c6fac17726-Paper-Conference.pdf},
  volume = {2025},
  year = {2025}
}

@misc{yang2025qwen3,
      title={Qwen3 Technical Report}, 
      author={An Yang and Anfeng Li and Baosong Yang and Beichen Zhang and Binyuan Hui and Bo Zheng and Bowen Yu and Chang Gao and Chengen Huang and Chenxu Lv and Chujie Zheng and Dayiheng Liu and Fan Zhou and Fei Huang and Feng Hu and Hao Ge and Haoran Wei and Huan Lin and Jialong Tang and Jian Yang and Jianhong Tu and Jianwei Zhang and Jianxin Yang and Jiaxi Yang and Jing Zhou and Jingren Zhou and Junyang Lin and Kai Dang and Keqin Bao and Kexin Yang and Le Yu and Lianghao Deng and Mei Li and Mingfeng Xue and Mingze Li and Pei Zhang and Peng Wang and Qin Zhu and Rui Men and Ruize Gao and Shixuan Liu and Shuang Luo and Tianhao Li and Tianyi Tang and Wenbiao Yin and Xingzhang Ren and Xinyu Wang and Xinyu Zhang and Xuancheng Ren and Yang Fan and Yang Su and Yichang Zhang and Yinger Zhang and Yu Wan and Yuqiong Liu and Zekun Wang and Zeyu Cui and Zhenru Zhang and Zhipeng Zhou and Zihan Qiu},
      year={2025},
      eprint={2505.09388},
      archivePrefix={arXiv},
      primaryClass={cs.CL},
      url={https://arxiv.org/abs/2505.09388}, 
}

@misc{yang2025qwen25,
      title={Qwen2.5 Technical Report}, 
      author={An Yang and Baosong Yang and Beichen Zhang and Binyuan Hui and Bo Zheng and Bowen Yu and Chengyuan Li and Dayiheng Liu and Fei Huang and Haoran Wei and Huan Lin and Jian Yang and Jianhong Tu and Jianwei Zhang and Jianxin Yang and Jiaxi Yang and Jingren Zhou and Junyang Lin and Kai Dang and Keming Lu and Keqin Bao and Kexin Yang and Le Yu and Mei Li and Mingfeng Xue and Pei Zhang and Qin Zhu and Rui Men and Runji Lin and Tianhao Li and Tianyi Tang and Tingyu Xia and Xingzhang Ren and Xuancheng Ren and Yang Fan and Yang Su and Yichang Zhang and Yu Wan and Yuqiong Liu and Zeyu Cui and Zhenru Zhang and Zihan Qiu},
      year={2025},
      eprint={2412.15115},
      archivePrefix={arXiv},
      primaryClass={cs.CL},
      url={https://arxiv.org/abs/2412.15115}, 
}

@misc{openai_models,
  author       = {{OpenAI}},
  title        = {OpenAI API Models},
  year         = {2026},
  howpublished = {\url{https://developers.openai.com/api/docs/models}},
  note         = {Accessed 2026}
}

@article{robertson2009probabilistic,
author = {Robertson, Stephen and Zaragoza, Hugo},
title = {The Probabilistic Relevance Framework: BM25 and Beyond},
year = {2009},
issue_date = {April 2009},
publisher = {Now Publishers Inc.},
address = {Hanover, MA, USA},
volume = {3},
number = {4},
issn = {1554-0669},
url = {https://doi.org/10.1561/1500000019},
doi = {10.1561/1500000019},
journal = {Found. Trends Inf. Retr.},
month = apr,
pages = {333–389},
numpages = {57}
}

@inproceedings{reimers-2019-sentence,
    title = "Sentence-{BERT}: Sentence Embeddings using {S}iamese {BERT}-Networks",
    author = "Reimers, Nils  and
      Gurevych, Iryna",
    editor = "Inui, Kentaro  and
      Jiang, Jing  and
      Ng, Vincent  and
      Wan, Xiaojun",
    booktitle = "Proceedings of the 2019 Conference on Empirical Methods in Natural Language Processing and the 9th International Joint Conference on Natural Language Processing (EMNLP-IJCNLP)",
    month = nov,
    year = "2019",
    address = "Hong Kong, China",
    publisher = "Association for Computational Linguistics",
    url = "https://aclanthology.org/D19-1410/",
    doi = "10.18653/v1/D19-1410",
    pages = "3982--3992"
}

@misc{openai2024embeddings,
  author       = {{OpenAI}},
  title        = {New Embedding Models and API Updates},
  year         = {2024},
  month        = jan,
  howpublished = {\url{https://openai.com/index/new-embedding-models-and-api-updates/}}
}

@misc{chhikara2025mem0,
      title={Mem0: Building Production-Ready AI Agents with Scalable Long-Term Memory}, 
      author={Prateek Chhikara and Dev Khant and Saket Aryan and Taranjeet Singh and Deshraj Yadav},
      year={2025},
      eprint={2504.19413},
      archivePrefix={arXiv},
      primaryClass={cs.CL},
      url={https://arxiv.org/abs/2504.19413}, 
}

@inproceedings{xu2026mem,
  author = {Xu, Wujiang and Liang, Zujie and Mei, Kai and Gao, Hang and Tan, Juntao and Zhang, Yongfeng},
  booktitle = {Advances in Neural Information Processing Systems},
  doi = {10.52202/085713-0593},
  editor = {D. Belgrave and C. Zhang and H. Lin and R. Pascanu and P. Koniusz and M. Ghassemi and N. Chen},
  pages = {17577--17604},
  publisher = {Curran Associates, Inc.},
  title = {A-Mem: Agentic Memory for LLM Agents},
  url = {https://proceedings.neurips.cc/paper_files/paper/2025/file/19909c36f51abc4856b4560aff3d36d6-Paper-Conference.pdf},
  volume = {38},
  year = {2025}
}

@misc{li2025memos_long,
      title={MemOS: A Memory OS for AI System}, 
      author={Zhiyu Li and Chenyang Xi and Chunyu Li and Ding Chen and Boyu Chen and Shichao Song and Simin Niu and Hanyu Wang and Jiawei Yang and Chen Tang and Qingchen Yu and Jihao Zhao and Yezhaohui Wang and Peng Liu and Zehao Lin and Pengyuan Wang and Jiahao Huo and Tianyi Chen and Kai Chen and Kehang Li and Zhen Tao and Huayi Lai and Hao Wu and Bo Tang and Zhengren Wang and Zhaoxin Fan and Ningyu Zhang and Linfeng Zhang and Junchi Yan and Mingchuan Yang and Tong Xu and Wei Xu and Huajun Chen and Haofen Wang and Hongkang Yang and Wentao Zhang and Zhi-Qin John Xu and Siheng Chen and Feiyu Xiong},
      year={2025},
      eprint={2507.03724},
      archivePrefix={arXiv},
      primaryClass={cs.CL},
      url={https://arxiv.org/abs/2507.03724}, 
}

@misc{guo2026tmem,
  title         = {T-Mem: Memory That Anticipates, Not Archives},
  author        = {Guo, Weidong and Wang, Dakai and Wang, Zixuan and Liu, Hui and Xu, Yu},
  year          = {2026},
  eprint        = {2606.15405},
  archivePrefix = {arXiv},
  primaryClass  = {cs.CL},
  note          = {Accepted to EMNLP 2026 Main Conference},
  url           = {https://arxiv.org/abs/2606.15405},
}

@inproceedings{zheng2023judging,
  author = {Zheng, Lianmin and Chiang, Wei-Lin and Sheng, Ying and Zhuang, Siyuan and Wu, Zhanghao and Zhuang, Yonghao and Lin, Zi and Li, Zhuohan and Li, Dacheng and Xing, Eric and Zhang, Hao and Gonzalez, Joseph and Stoica, Ion},
  booktitle = {Advances in Neural Information Processing Systems},
  doi = {10.52202/075280-2020},
  editor = {A. Oh and T. Naumann and A. Globerson and K. Saenko and M. Hardt and S. Levine},
  pages = {46595--46623},
  publisher = {Curran Associates, Inc.},
  title = {Judging LLM-as-a-Judge with MT-Bench and Chatbot Arena},
  url = {https://proceedings.neurips.cc/paper_files/paper/2023/file/91f18a1287b398d378ef22505bf41832-Paper-Datasets_and_Benchmarks.pdf},
  volume = {36},
  year = {2023}
}

@misc{li2024llmsasjudges,
      title={LLMs-as-Judges: A Comprehensive Survey on LLM-based Evaluation Methods}, 
      author={Haitao Li and Qian Dong and Junjie Chen and Huixue Su and Yujia Zhou and Qingyao Ai and Ziyi Ye and Yiqun Liu},
      year={2024},
      eprint={2412.05579},
      archivePrefix={arXiv},
      primaryClass={cs.CL},
      url={https://arxiv.org/abs/2412.05579}, 
}

@article{einstein2005multiple,
 ISSN = {09637214},
 URL = {http://www.jstor.org/stable/20183049},
 author = {Gilles O. Einstein and Mark A. McDaniel},
 journal = {Current Directions in Psychological Science},
 number = {6},
 pages = {286--290},
 publisher = {[Association for Psychological Science, Sage Publications, Inc.]},
 title = {Prospective Memory: Multiple Retrieval Processes},
 urldate = {2026-10-03},
 volume = {14},
 year = {2005}
}

@article{einstein1990normal,
  author    = {Einstein, Gilles O. and McDaniel, Mark A.},
  title     = {Normal Aging and Prospective Memory},
  journal   = {Journal of Experimental Psychology: Learning, Memory, and Cognition},
  year      = {1990},
  volume    = {16},
  number    = {4},
  pages     = {717--726},
  publisher = {American Psychological Association},
  doi       = {10.1037/0278-7393.16.4.717},
  issn      = {1939-1285, 0278-7393},
  pmid      = {2142956}
}

@article{radvansky2011event,
  title={Event perception},
  author={Radvansky, Gabriel A and Zacks, Jeffrey M},
  journal={Wiley Interdisciplinary Reviews: Cognitive Science},
  volume={2},
  number={6},
  pages={608--620},
  year={2011},
  publisher={Wiley Online Library}
}

@article{McDaniel2000S127,
  title={Strategic and automatic processes in prospective memory retrieval: A multiprocess framework},
  author={McDaniel, Mark A and Einstein, Gilles O},
  journal={Applied Cognitive Psychology: The Official Journal of the Society for Applied Research in Memory and Cognition},
  volume={14},
  number={7},
  pages={S127--S144},
  year={2000},
  publisher={Wiley Online Library}
}

@inproceedings{
hu2022lora,
title={Lo{RA}: Low-Rank Adaptation of Large Language Models},
author={Edward J Hu and Yelong Shen and Phillip Wallis and Zeyuan Allen-Zhu and Yuanzhi Li and Shean Wang and Lu Wang and Weizhu Chen},
booktitle={International Conference on Learning Representations},
year={2022},
url={https://openreview.net/forum?id=nZeVKeeFYf9}
}

@inproceedings{mittal-2026-forget,
    title = "Did You Forget What {I} Asked? Prospective Memory Failures in Large Language Models",
    author = "Mittal, Avni",
    editor = "Chang, Kai-Wei  and
      Mehrabi, Ninareh  and
      Krishna, Satyapriya  and
      Das, Anubrata  and
      Dhamala, Jwala  and
      Cao, Yang Trista  and
      Kumarage, Tharindu  and
      Ramakrishna, Anil  and
      Christodoulopoulos, Christos  and
      Wan, Yixin  and
      Galystan, Aram  and
      Kumar, Anoop  and
      Gupta, Rahul",
    booktitle = "Proceedings of the 6th Workshop on Trustworthy {NLP} ({T}rust{NLP} 2026)",
    month = jul,
    year = "2026",
    address = "San Diego, California",
    publisher = "Association for Computational Linguistics",
    url = "https://aclanthology.org/2026.trustnlp-main.33/",
    doi = "10.18653/v1/2026.trustnlp-main.33",
    pages = "471--488",
    ISBN = "979-8-89176-418-7"
}

@article{zhang2026triggerbench,
  title={TriggerBench: Investigating Prospective Memory for Large Language Models},
  author={Zhang, Tianhua and Wang, Xinjiang and Zhang, Qianxi and Chen, Qi and Li, Kun and Chen, Yaoqi and Wang, Dingdong and Meng, Helen and Lu, Yan},
  journal={arXiv preprint arXiv:2606.23459},
  year={2026}
}

@inproceedings{liu2026pm,
  title={PM-Bench: Evaluating Prospective Memory of LLM Agents},
  author={Liu, Genglin and Gabriel, Saadia},
  booktitle={Third Conference on Language Modeling},
  year={2026}
}

@article{tang2026proagentbench,
  title={Proagentbench: Evaluating llm agents for proactive assistance with real-world data},
  author={Tang, Yuanbo and Tang, Huaze and Cao, Tingyu and Nguyen, Lam and Zhang, Anping and Cao, Xinwen and Liu, Chunkang and Ding, Wenbo and Li, Yang},
  journal={arXiv preprint arXiv:2602.04482},
  year={2026}
}

@inproceedings{
yang2026fingertip,
title={FingerTip 20K: A Benchmark for Proactive and Personalized Mobile {LLM} Agents},
author={Qinglong Yang and Haoming Li and Haotian Zhao and Xiaokai Yan and Jingtao Ding and Fengli Xu and Yong Li},
booktitle={The Fourteenth International Conference on Learning Representations},
year={2026},
url={https://openreview.net/forum?id=n3iFV0gLMc}
}

@misc{yang2026proagent,
      title={ProAgent: Harnessing On-Demand Sensory Contexts for Proactive LLM Agent Systems in the Wild}, 
      author={Bufang Yang and Lilin Xu and Liekang Zeng and Yunqi Guo and Siyang Jiang and Wenrui Lu and Kaiwei Liu and Yixuan Li and Xiaofan Jiang and Guoliang Xing and Zhenyu Yan},
      year={2026},
      eprint={2512.06721},
      archivePrefix={arXiv},
      primaryClass={cs.AI},
      url={https://arxiv.org/abs/2512.06721}, 
}

@inproceedings{li-etal-2026-locomo,
    title = "Locomo-Plus: Beyond-Factual Cognitive Memory Evaluation Framework for {LLM} Agents",
    author = "Li, Yifei  and
      Guo, Weidong  and
      Zhang, Lingling  and
      Xu, Rongman  and
      Huang, Muye  and
      Liu, Hui  and
      Xu, Lijiao  and
      Xu, Yu  and
      Liu, Jun",
    editor = "Liakata, Maria  and
      Moreira, Viviane P.  and
      Zhang, Jiajun  and
      Jurgens, David",
    booktitle = "Proceedings of the 64th Annual Meeting of the {A}ssociation for {C}omputational {L}inguistics (Volume 1: Long Papers)",
    month = jul,
    year = "2026",
    address = "San Diego, California, United States",
    publisher = "Association for Computational Linguistics",
    url = "https://aclanthology.org/2026.acl-long.1150/",
    doi = "10.18653/v1/2026.acl-long.1150",
    pages = "25085--25100",
    ISBN = "979-8-89176-390-6"
}

@inproceedings{
gutierrez2025from,
title={From {RAG} to Memory: Non-Parametric Continual Learning for Large Language Models},
author={Bernal Jim{\'e}nez Guti{\'e}rrez and Yiheng Shu and Weijian Qi and Sizhe Zhou and Yu Su},
booktitle={Forty-second International Conference on Machine Learning},
year={2025},
url={https://openreview.net/forum?id=LWH8yn4HS2}
}

@inproceedings{yue-etal-2026-hypermem,
    title = "{H}yper{M}em: Hypergraph Memory for Long-Term Conversations",
    author = "Yue, Juwei  and
      Hu, Chuanrui  and
      Sheng, Jiawei  and
      Zhou, Zuyi  and
      Zhang, Wenyuan  and
      Liu, Tingwen  and
      Guo, Li  and
      Deng, Yafeng",
    editor = "Liakata, Maria  and
      Moreira, Viviane P.  and
      Zhang, Jiajun  and
      Jurgens, David",
    booktitle = "Proceedings of the 64th Annual Meeting of the {A}ssociation for {C}omputational {L}inguistics (Volume 1: Long Papers)",
    month = jul,
    year = "2026",
    address = "San Diego, California, United States",
    publisher = "Association for Computational Linguistics",
    url = "https://aclanthology.org/2026.acl-long.1627/",
    doi = "10.18653/v1/2026.acl-long.1627",
    pages = "35237--35254",
    ISBN = "979-8-89176-390-6"
}

@inproceedings{kang2025memoryos,
    title = "Memory {OS} of {AI} Agent",
    author = "Kang, Jiazheng  and
      Ji, Mingming  and
      Zhao, Zhe  and
      Bai, Ting",
    editor = "Christodoulopoulos, Christos  and
      Chakraborty, Tanmoy  and
      Rose, Carolyn  and
      Peng, Violet",
    booktitle = "Proceedings of the 2025 Conference on Empirical Methods in Natural Language Processing",
    month = nov,
    year = "2025",
    address = "Suzhou, China",
    publisher = "Association for Computational Linguistics",
    url = "https://aclanthology.org/2025.emnlp-main.1318/",
    doi = "10.18653/v1/2025.emnlp-main.1318",
    pages = "25961--25970",
    ISBN = "979-8-89176-332-6"
}

@article{packer2023memgpt,
  title={Memgpt: Towards llms as operating systems},
  author={Packer, Charles and Wooders, Sarah and Lin, Kevin and Fang, Vivian and Patil, Shishir G and Stoica, Ion and Gonzalez, Joseph E},
  journal={arXiv preprint arXiv:2310.08560},
  year={2023}
}

@article{bain2022whisperx,
  title={WhisperX: Time-Accurate Speech Transcription of Long-Form Audio},
  author={Bain, Max and Huh, Jaesung and Han, Tengda and Zisserman, Andrew},
  journal={INTERSPEECH 2023},
  year={2023}
}

@misc{bai2025qwen3vltechnicalreport,
      title={Qwen3-VL Technical Report}, 
      author={Shuai Bai and Yuxuan Cai and Ruizhe Chen and Keqin Chen and Xionghui Chen and Zesen Cheng and Lianghao Deng and Wei Ding and Chang Gao and Chunjiang Ge and Wenbin Ge and Zhifang Guo and Qidong Huang and Jie Huang and Fei Huang and Binyuan Hui and Shutong Jiang and Zhaohai Li and Mingsheng Li and Mei Li and Kaixin Li and Zicheng Lin and Junyang Lin and Xuejing Liu and Jiawei Liu and Chenglong Liu and Yang Liu and Dayiheng Liu and Shixuan Liu and Dunjie Lu and Ruilin Luo and Chenxu Lv and Rui Men and Lingchen Meng and Xuancheng Ren and Xingzhang Ren and Sibo Song and Yuchong Sun and Jun Tang and Jianhong Tu and Jianqiang Wan and Peng Wang and Pengfei Wang and Qiuyue Wang and Yuxuan Wang and Tianbao Xie and Yiheng Xu and Haiyang Xu and Jin Xu and Zhibo Yang and Mingkun Yang and Jianxin Yang and An Yang and Bowen Yu and Fei Zhang and Hang Zhang and Xi Zhang and Bo Zheng and Humen Zhong and Jingren Zhou and Fan Zhou and Jing Zhou and Yuanzhi Zhu and Ke Zhu},
      year={2025},
      eprint={2511.21631},
      archivePrefix={arXiv},
      primaryClass={cs.CV},
      url={https://arxiv.org/abs/2511.21631}, 
}

@article{rasmussen2025zep,
  title={Zep: a temporal knowledge graph architecture for agent memory},
  author={Rasmussen, Preston and Paliychuk, Pavlo and Beauvais, Travis and Ryan, Jack and Chalef, Daniel},
  journal={arXiv preprint arXiv:2501.13956},
  year={2025}
}

@inproceedings{zhong2024memorybank,
  author    = {Wanjun Zhong and Lianghong Guo and Qiqi Gao and He Ye and Yanlin Wang},
  title     = {{MemoryBank}: Enhancing Large Language Models with Long-Term Memory},
  booktitle = {Proceedings of the AAAI Conference on Artificial Intelligence},
  volume    = {38},
  number    = {17},
  pages     = {19724--19731},
  year      = {2024},
  doi       = {10.1609/aaai.v38i17.29946},
  url       = {https://ojs.aaai.org/index.php/AAAI/article/view/29946}
}

@misc{meta_rayban,
  author       = {{Meta}},
  title        = {{Ray-Ban Meta} {AI} Glasses},
  year         = {2026},
  howpublished = {\url{https://www.meta.com/ai-glasses/ray-ban-meta/}},
  note         = {Accessed: 2026-10-10}
}

@article{hurst2024gpt,
  title={Gpt-4o system card},
  author={Aaron Hurst and Adam Lerer and Adam P. Goucher and Adam Perelman and Aditya Ramesh and Aidan Clark and AJ Ostrow and Akila Welihinda and Alan Hayes and Alec Radford and others},
  journal={arXiv preprint arXiv:2410.21276},
  year={2024}
}

@misc{anthropic2026sonnet46,
  author       = {{Anthropic}},
  title        = {System Card: {Claude Sonnet 4.6}},
  year         = {2026},
  howpublished = {\url{https://www.anthropic.com/claude-sonnet-4-6-system-card}}
}

@misc{comanici2025gemini,
      title={Gemini 2.5: Pushing the Frontier with Advanced Reasoning, Multimodality, Long Context, and Next Generation Agentic Capabilities},
      author={Gheorghe Comanici and Eric Bieber and Mike Schaekermann and Ice Pasupat and Noveen Sachdeva and Inderjit Dhillon and Marcel Blistein and Ori Ram and Dan Zhang and Evan Rosen and others},
      year={2025},
      eprint={2507.06261},
      archivePrefix={arXiv},
      primaryClass={cs.CL},
      url={https://arxiv.org/abs/2507.06261}, 
}

@inproceedings{lyu-etal-2026-personalalign,
    title = "{P}ersonal{A}lign: Hierarchical Implicit Intent Alignment for Personalized {GUI} Agent with Long-Term User-Centric Records",
    author = "Lyu, Yibo  and
      Chen, Gongwei  and
      Shao, Rui  and
      Guan, Weili  and
      Nie, Liqiang",
    editor = "Liakata, Maria  and
      Moreira, Viviane P.  and
      Zhang, Jiajun  and
      Jurgens, David",
    booktitle = "Proceedings of the 64th Annual Meeting of the {A}ssociation for {C}omputational {L}inguistics (Volume 1: Long Papers)",
    month = jul,
    year = "2026",
    address = "San Diego, California, United States",
    publisher = "Association for Computational Linguistics",
    url = "https://aclanthology.org/2026.acl-long.1669/",
    doi = "10.18653/v1/2026.acl-long.1669",
    pages = "36074--36089",
    ISBN = "979-8-89176-390-6"
}

@article{scullin2009evidence,
  title={Evidence for spontaneous retrieval of suspended but not finished prospective memories},
  author={Scullin, Michael K. and Einstein, Gilles O. and McDaniel, Mark A.},
  journal={Memory \& Cognition},
  volume={37},
  number={4},
  pages={425--433},
  year={2009},
  doi={10.3758/MC.37.4.425}
}

@misc{yang2026sensorpersona,
      title={SensorPersona: An LLM-Empowered System for Continual Persona Extraction from Longitudinal Mobile Sensor Streams}, 
      author={Bufang Yang and Lilin Xu and Yixuan Li and Kaiwei Liu and Xiaofan Jiang and Zhenyu Yan},
      year={2026},
      eprint={2604.06204},
      archivePrefix={arXiv},
      primaryClass={cs.CL},
      url={https://arxiv.org/abs/2604.06204}, 
}

@misc{meta2026muse,
  author       = {{Meta}},
  title        = {Introducing {Muse}: The World's First Personal {AI} Agent Built for Everyone},
  howpublished = {\url{https://about.fb.com/news/2026/09/introducing-muse-personal-ai-agent/}},
  year         = {2026},
  month        = sep,
  note         = {Accessed: 2026-10-12}
}

@misc{openai2026dots,
  author       = {{OpenAI}},
  title        = {Introducing dots},
  howpublished = {\url{https://openai.com/index/introducing-dots/}},
  year         = {2026},
  month        = sep,
  note         = {Accessed: 2026-10-12}
}

@misc{meta2024llama33,
  author       = {{Meta AI}},
  title        = {{Llama 3.3 70B Instruct}},
  year         = {2024},
  howpublished = {\url{https://huggingface.co/meta-llama/Llama-3.3-70B-Instruct}},
  note         = {Hugging Face model card}
}

@techreport{anthropic2026claudeopus46,
  author      = {{Anthropic}},
  title       = {{Claude Opus 4.6 System Card}},
  institution = {Anthropic},
  year        = {2026},
  url         = {https://www-cdn.anthropic.com/c788cbc0a3da9135112f97cdf6dcd06f2c16cee2.pdf}
}

\appendix
\definecolor{promptframe}{HTML}{4D4D4D}
\definecolor{promptback}{HTML}{FCFCFC}   %
\definecolor{promptvar}{HTML}{1F5BB5}   %
\newtcolorbox{promptbox}[2][]{enhanced, breakable=false, colback=promptback, colframe=promptframe, coltitle=white,
  fonttitle=\bfseries\sffamily\footnotesize, title={#2}, halign title=center, arc=1.2mm, boxrule=0.8pt,
  left=6pt, right=6pt, top=3pt, bottom=4pt, fontupper=\sffamily\scriptsize, #1}   %
\definecolor{promptlabel}{HTML}{E4E4E4}
\newcommand{\psec}[1]{\par\smallskip\noindent{\setlength{\fboxsep}{1.5pt}\colorbox{promptlabel}{\bfseries #1}}\hspace{0.5em}\ignorespaces}   %
\newcommand{\psep}{\par\smallskip\noindent{\color{promptframe!70}\rule{\linewidth}{0.4pt}}\par}  %
\newcommand{\pvar}[1]{{\color{promptvar}\ttfamily\{#1\}}}                                  %
\newcommand{\pfill}{{\color{gray}\itshape [prompt text to be filled]}}                      %

\section{Dataset Details}
\label{sec:appendix_dataset}

\subsection{Instance Examples}
\label{sec:appendix_examples}
\definecolor{exform}{HTML}{1F5FA8}
\definecolor{exdec}{HTML}{B35900} 
\definecolor{exsc}{HTML}{2E7D32}  
\definecolor{exmod}{HTML}{7B1FA2} 
\newcommand{\exev}[2]{{\ttfamily #1} \textbar{} vision: #2}         
\newcommand{\exch}[2]{ \textbar{} #1: #2}                            
\newcommand{\exgap}[1]{{\color{gray}\itshape [\,#1\,]}}  
\newcommand{\exlab}[2]{{\color{#1}\bfseries #2}}         
\definecolor{exframe}{HTML}{1F3A5F}
\definecolor{exback}{HTML}{FFFFFF}    %
\definecolor{exlabel}{HTML}{D6E0EE}   %
\newtcolorbox{examplebox}[1]{enhanced, breakable=false, colback=exback, colframe=exframe, coltitle=white,
  fonttitle=\bfseries\sffamily\footnotesize, title={#1}, halign title=center, arc=1.2mm, boxrule=0.8pt,
  left=6pt, right=6pt, top=3pt, bottom=4pt, fontupper=\sffamily\scriptsize}
\newcommand{\esec}[1]{\par\smallskip\noindent{\setlength{\fboxsep}{1.5pt}\colorbox{exlabel}{\bfseries #1}}\hspace{0.5em}\ignorespaces}
\newcommand{\esep}{\par\smallskip\noindent{\color{exframe!60}\rule{\linewidth}{0.4pt}}\par}
\newcommand{\exvartab}[1]{{\renewcommand{\arraystretch}{1.15}\setlength{\tabcolsep}{4pt}%
\begin{tabular}{@{}p{1.6cm}p{10.6cm}p{2.5cm}@{}}
\toprule
\textbf{Variant} & \textbf{Key event} & \textbf{Decision at $e_d$} \\
#1
\bottomrule
\end{tabular}}}

\begin{figure*}[t]
\centering
\begin{examplebox}{Instance Example: One Scenario and Its Four Variants}
\esec{Personal Context}
Lives alone in a small apartment. Works at a downtown office.
\esep
\esec{Timeline}
{\raggedright
\exev{Fri 2026-07-03 07:33}{The user taps the tablet screen propped against a salt shaker, scrolling a page of densely spaced text with a narrow sidebar of grey icons. A half-eaten piece of toast rests on a yellow plate to the left.}\par
\exgap{53 daily-life events, e.g., office work, commuting, reading at home}\par
{\color{exform}\exev{Mon 2026-07-06 07:50}{The user stands at the bathroom sink and squeezes a flattened toothpaste tube over the toothbrush, rolling it tightly from the bottom. Only a tiny smear reaches the bristles. After squeezing it once more with no visible result, the user drops the tube into the trash bin beside the sink.}}\hfill\exlab{exform}{$\leftarrow e_f$}\par
\exgap{143 daily-life events}\par
\exev{Tue 2026-07-07 18:05}{The user stands in a crowded subway car gripping a metal pole, a folded newspaper under one arm, as station lights and tiled walls streak past the dark window.}\exch{audio}{Next stop, Central Station. Doors will open on the left.}\par
{\color{exdec}\exev{Tue 2026-07-07 18:35}{The user walks into a supermarket through the automatic doors and picks up a shopping basket from the stack by the entrance. Aisles of packaged food stretch ahead under bright ceiling lights.}}\hfill\exlab{exdec}{$\leftarrow e_d$ (current moment)}\par}
\esep
\esec{Annotation}
Delayed intention: Buy a new tube of toothpaste the next time the user is at a store that sells toiletries.
Decision: assist.
\esep
\esec{Other Variants}
Each variant changes one event of the \texttt{Positive} timeline (key event); all other events are identical.\par\smallskip
\exvartab{%
\midrule
\texttt{Negative} & \exlab{exdec}{$e_d$} \exev{Tue 2026-07-07 18:35}{The user steps into a small perfume shop, where illuminated shelves display glass fragrance bottles and boxed gift sets. A staff member rearranges sample strips on a counter near the entrance.} & Stay silent: a perfume shop does not sell toothpaste. \\
\midrule
\texttt{State Change} & \exlab{exsc}{$e_{sc}$} \exev{Mon 2026-07-06 08:05}{The user opens the cabinet under the sink, reaches inside, and takes out an unopened tube of toothpaste. The user places it beside the toothbrush cup on the sink.} & Stay silent: a new tube is already at home. \\
\midrule
\texttt{No Formation} & \exlab{exform}{$e_f$} \exev{Mon 2026-07-06 07:50}{The user crouches to lace brown leather shoes by the front door, tugging each knot tight, then stands and slips a slim wallet and a ring of keys into a jacket pocket.} & Stay silent: no intention is formed. \\}
\end{examplebox}
\caption{An example scenario from \benchmark and its four variants, showing representative events from the \texttt{Positive} timeline and its annotation, followed by the variant-specific events of the other three variants.}
\label{fig:example_timeline}
\end{figure*}

Fig.~\ref{fig:example_timeline} presents an example scenario and its four variants, including \texttt{Positive}, \texttt{Negative}, \texttt{State Change} and \texttt{No Formation}.
Each instance includes personal context and a timeline of $N=200$ timestamped events, each containing sensory context.
The hidden gold annotation specifies the delayed intention and the expected proactive behavior.
The \texttt{Negative}, \texttt{State Change}, and \texttt{No Formation} variants are then shown through their variant-specific event, while all remaining events are shared with the \texttt{Positive} timeline.

\subsection{Benchmark Construction}
\label{sec:appendix_construction}

\noindent\textbf{Seed Scenario Construction.}
Three annotators first create seed scenarios drawn from diverse everyday experiences.
Each scenario contains one or more implicit delayed intentions, together with their formation and decision moments, relevant personal context, and textual descriptions of the visual, speech, and notification context at each key moment.
Annotators additionally specify the variant-specific events required to construct the four controlled variants and provide reasoning traces explaining whether proactive assistance should be provided or withheld.
All seed scenarios are cross-reviewed by another annotator.

\noindent\textbf{Scenario Diversification and Validation.}
Starting from seed scenarios, we then use LLMs with the prompts in Fig.~\ref{fig:prompt_scenario} to generate a larger and more diverse set of candidate scenarios across everyday categories.
The generated scenarios are subsequently reviewed by human annotators to remove implausible or redundant cases and verify the consistency of the annotated intentions and variants.

\noindent\textbf{Timeline Construction.}
For each validated scenario, we construct four timeline variants with a shared temporal structure.
We first build an initial timeline containing the variant-specific events and additional scenario-specific daily-life events generated consistently with the personal context.
These additional events are intention-independent and serve as distractors between the key moments of the scenario.
The formation moment $e_f$ is placed earlier in the timeline, while the shared decision moment $e_d$ is the final event.
We further augment each timeline with a pool of shared daily-life background events, inserted both before and between the scenario-specific events.
Unlike the scenario-specific distractors, these background events are reused across scenarios, with their timestamps adjusted to fit each timeline.
Both types of distractor events are designed to be independent of the target intention and to neither reveal nor resolve it.
Within a scenario, all non-variant events are shared across the four variants so that they differ only in their controlled variant-specific events.
Each resulting timeline contains $N=200$ events.

\noindent\textbf{Sensor Data Pairing.}
For \benchmarklite, we pair events represented by textual descriptions with egocentric video and audio.
We first retrieve video clips from S-EMBER~\cite{wang2026sember} using the visual descriptions in each event and synthesize the corresponding audio from its textual speech content using text-to-speech~\cite{azure_tts}.
S-EMBER is released under the CC BY-NC 4.0 license, and our use of the dataset complies with its terms.
We record the original video identifier and the start and end timestamps of each clip, so that every clip can be traced back to its source in S-EMBER.
When retrieval cannot reliably match the visual description or recorded audio is required, particularly for key formation and decision moments, we self-record additional egocentric video and audio.
Both retrieved and self-recorded videos were captured using Meta smart glasses~\cite{meta_rayban}.
All paired video and audio are verified by human annotators against the corresponding event descriptions.
Our self-recorded sensor collection was approved by the institutional IRB, and all participants provided informed consent prior to data collection. Faces and other visually identifying or sensitive regions were blurred to protect participant privacy.
Speech in the recordings followed scripted scenarios with synthetic names and contains no personal information about participants.

\section{Implementation Details}
\label{sec:appendix_implementation}

\noindent\textbf{Default Setup.}
We conduct all experiments on a server with two NVIDIA RTX PRO 6000 GPUs.
We randomly split \benchmark~into 60\% training and 40\% testing, assigning all variants from the same scenario to the same split.
Unless otherwise specified, \workname~and fine-tuned proactive-agent baselines use Qwen3-8B with LoRA of rank 8 and are trained for five epochs using AdamW with a learning rate of $5\times10^{-4}$ and 10\% warmup, with a fixed random seed for all experiments for fair comparison.
Claude-Opus-4.6~\cite{anthropic2026claudeopus46} is used to generate reasoning traces for distillation.
For LLM-based memory construction, we use GPT-4o-mini and text-embedding-3-small for embedding-based similarity.
RAG-based and memory-system baselines use GPT-5.4 for proactive reasoning.
Unless otherwise specified, the reasoner receives the personal context and current event together with the historical context provided by each method.
For methods using top-$k$ retrieval, we set $k=10$ unless otherwise specified.

\noindent\textbf{\workname.}
For context extraction, \workname~employs Qwen3-VL-32B~\cite{bai2025qwen3vltechnicalreport} to caption each egocentric video clip using the prompt in Fig.~\ref{fig:prompt_caption}, and WhisperX~\cite{bain2022whisperx} to transcribe speech with speaker diarization.
For cross-event linking in memory construction, \workname~considers the top-$m$ candidate facts with $m=8$ for each new event.
At the decision moment, it retrieves $k=10$ past events, with $k^C=2$ for content retrieval and $k^{CT}=8$ for cue-triggered retrieval.
For reasoner training, each historical context contains $K\in\{8,10,12\}$ past events from the initial timeline.
For \texttt{Positive}, \texttt{Negative}, and \texttt{State Change} instances, the formation moment $e_f$ is always included, while \texttt{State Change} instances additionally include $e_{sc}$.
The remaining events are sampled at random.

\noindent\textbf{Baseline Implementations.}
For Proactive Agent~\cite{lu2025proactive} and ContextAgent~\cite{yang2025contextagent}, we follow their original input and prompting setups while adapting them to our task.
Proactive Agent receives the complete timeline as event history without personal context, and predicts whether to provide proactive assistance together with a response.
ContextAgent receives only the current sensory context and personal context, without access to past events.
We follow the original data-construction procedures of both methods under the default training setup above.
Full-context LLMs use the complete past timeline as historical context, with the default generation settings of each model.
RAG-based methods use the current event as the query and retrieve past events according to their respective lexical or embedding similarity.
For memory-system baselines, we use the official implementations of Mem0~\cite{chhikara2025mem0}, A-Mem~\cite{xu2026mem}, MemOS~\cite{li2025memos_long}, and T-Mem~\cite{guo2026tmem}.
Mem0 processes up to five consecutive same-day events per memory update, while A-Mem and MemOS process one event per update.
Except for T-Mem, all retrieval-based baselines use $k=10$ at their native retrieval granularity (e.g., raw events for RAG, facts for Mem0).
For T-Mem, we evaluate its default setting, which retrieves 5 scenes and 15 items, and a horizon-based variant that ranks scenes solely by similarity between the current event and each scene's horizon trigger and retrieves top-$10$ scenes. 
For a fair comparison, all full-context, RAG-based, and memory-system baselines pass their retrieved results to the reasoner together with the same personal context from the dataset.

\section{Prompts}
\label{sec:appendix_prompts}

This section presents the details of the prompts used in this study, including those for sensory context extraction, memory construction, and proactive reasoning in \workname, as well as prompts for the baselines, data generation, and the LLM judge.

\begin{figure}[t]
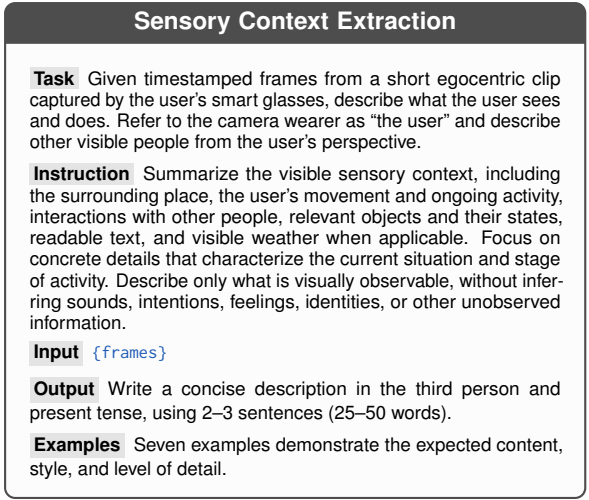

\centering
\begin{promptbox}{Sensory Context Extraction}
\psec{Task}
Given timestamped frames from a short egocentric clip captured by the user's smart glasses, describe what the user sees and does.
Refer to the camera wearer as ``the user'' and describe other visible people from the user's perspective.\par

\psec{Instruction}
Summarize the visible sensory context, including the surrounding place, the user's movement and ongoing activity, interactions with other people, relevant objects and their states, readable text, and visible weather when applicable.
Focus on concrete details that characterize the current situation and stage of activity.
Describe only what is visually observable, without inferring sounds, intentions, feelings, identities, or other unobserved information.\par

\psec{Input} \pvar{frames}\par
\psec{Output}
Write a concise description in the third person and present tense, using 2--3 sentences (25--50 words).\par

\psec{Examples}
Seven examples demonstrate the expected content, style, and level of detail.\par
\end{promptbox}

\caption{Summarized prompt for sensory context extraction in \workname.}
\label{fig:prompt_caption}
\end{figure}

\noindent\textbf{Sensory Context Extraction in \workname.}
Fig.~\ref{fig:prompt_caption} shows the summarized prompt used to extract visual context from egocentric video clips captured by smart glasses from the user's perspective.
The VLM is instructed to describe the surrounding environment, the user's movement and activity, and visible objects together with their states, providing the contextual elements needed to represent the observed event~\cite{radvansky2011event}.

\begin{figure}[t]
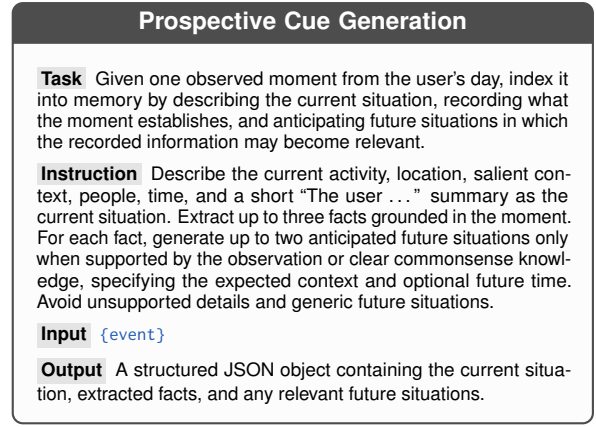

\centering
\begin{promptbox}{Prospective Cue Generation}
\psec{Task}
Given one observed moment from the user's day, index it into memory by describing the current situation, recording what the moment establishes, and anticipating future situations in which the recorded information may become relevant.\par

\psec{Instruction}
Describe the current activity, location, salient context, people, time, and a short ``The user \ldots'' summary as the current situation.
Extract up to three facts grounded in the moment.
For each fact, generate up to two anticipated future situations only when supported by the observation or clear commonsense knowledge, specifying the expected context and optional future time.
Avoid unsupported details and generic future situations.\par

\psec{Input} \pvar{event}\par

\psec{Output}
A structured JSON object containing the current situation, extracted facts, and any relevant future situations.\par
\end{promptbox}
\caption{Summarized prompt for prospective cue generation in \workname's memory construction.}
\label{fig:prompt_cue_generation}
\end{figure}

\begin{figure}[t]
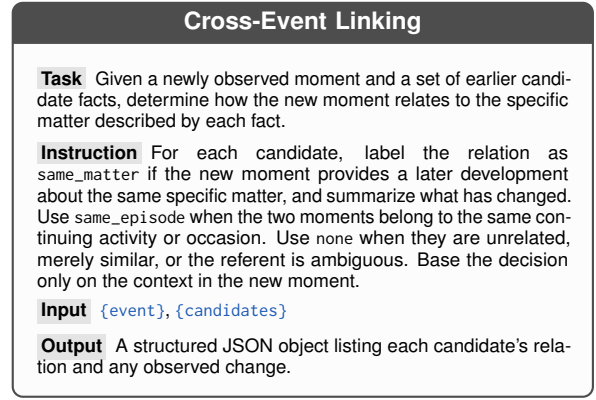

\centering
\begin{promptbox}{Cross-Event Linking}
\psec{Task}
Given a newly observed moment and a set of earlier candidate facts, determine how the new moment relates to the specific matter described by each fact.\par

\psec{Instruction}
For each candidate, label the relation as \texttt{same\_matter} if the new moment provides a later development about the same specific matter, and summarize what has changed.
Use \texttt{same\_episode} when the two moments belong to the same continuing activity or occasion.
Use \texttt{none} when they are unrelated, merely similar, or the referent is ambiguous.
Base the decision only on the context in the new moment.\par

\psec{Input} \pvar{event}, \pvar{candidates}\par

\psec{Output}
A structured JSON object listing each candidate's relation and any observed change.\par
\end{promptbox}
\caption{Summarized prompt for cross-event linking in \workname's memory construction.}
\label{fig:prompt_event_linking}
\end{figure}

\noindent\textbf{Memory Construction in \workname.}
Fig.~\ref{fig:prompt_cue_generation} and Fig.~\ref{fig:prompt_event_linking} show the prompts for prospective cue generation and cross-event linking.
Inspired by the event-based and time-based forms of prospective memory~\cite{einstein1990normal}, the cue-generation prompt instructs the memory-construction LLM to extract factual observations and anticipate observable contextual or temporal cues under which the specific factual observation may become useful.
The linking prompt instructs the LLM to determine whether a newly observed event is a subsequent development of an earlier fact in memory, enabling \workname~to capture how stored observations evolve over time, which is important for long-horizon proactive assistance.

\begin{figure}[t]
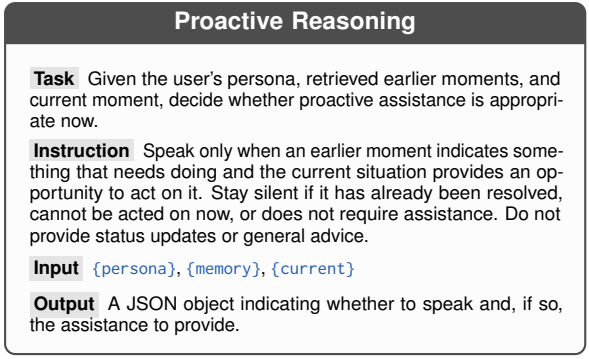

\centering
\begin{promptbox}{Proactive Reasoning}
\psec{Task}
Given the user's persona, retrieved earlier moments, and current moment, decide whether proactive assistance is appropriate now.\par

\psec{Instruction}
Speak only when an earlier moment indicates something that needs doing and the current situation provides an opportunity to act on it.
Stay silent if it has already been resolved, cannot be acted on now, or does not require assistance.
Do not provide status updates or general advice.\par

\psec{Input} \pvar{persona}, \pvar{memory}, \pvar{current}\par

\psec{Output}
A JSON object indicating whether to speak and, if so, the assistance to provide.\par
\end{promptbox}
\caption{Summarized prompt for proactive reasoning in \workname.}
\label{fig:prompt_reasoner}
\end{figure}

\noindent\textbf{Proactive Reasoning in \workname.}
Fig.~\ref{fig:prompt_reasoner} presents the prompt for proactive reasoning.
The prompt includes the personal context, retrieved past events in temporal order as historical context, and the current sensory context.
The task instructions guide the reasoner to determine whether the historical context provides evidence of an unresolved delayed intention and whether the current situation provides an appropriate opportunity to act on it.
The reasoner provides proactive assistance only when both conditions are satisfied. Otherwise, it remains silent.

\noindent\textbf{Prompts for Baselines.}
To ensure a fair comparison, we keep the task instructions and output format consistent with those used in \workname for full-context LLMs, retrieval-based methods, and memory systems used in our evaluation.
These baselines differ only in how historical information is provided to the reasoner, using either the full event history, the top-$k$ retrieved events, or the memories retrieved by the corresponding memory system.
For Proactive Agent~\cite{lu2025proactive} and ContextAgent~\cite{yang2025contextagent}, we follow their original prompting and input formulations to preserve their designed proactive reasoning settings.

\begin{figure}[t]
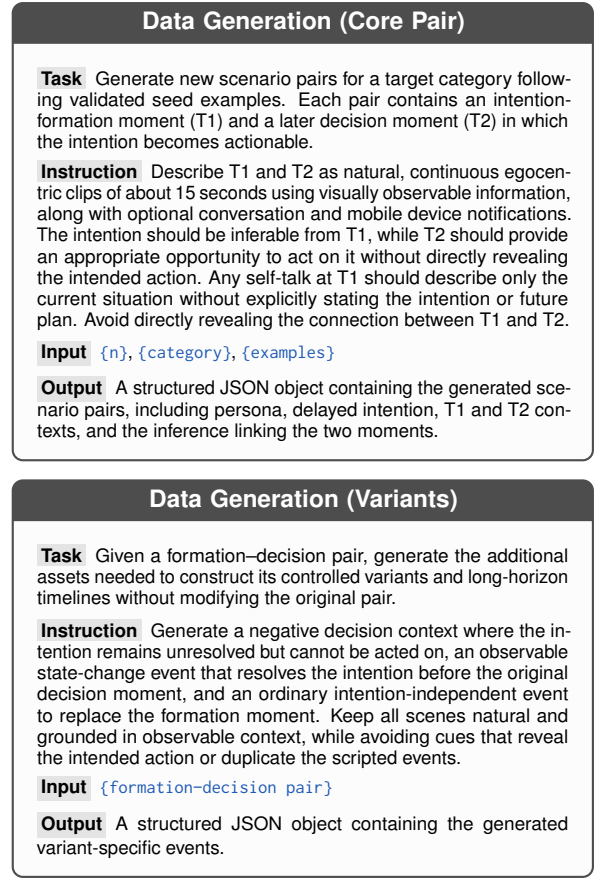

\centering
\begin{promptbox}{Data Generation (Core Pair)}
\psec{Task}
Generate new scenario pairs for a target category following validated seed examples.
Each pair contains an intention-formation moment (T1) and a later decision moment (T2) in which the intention becomes actionable.\par

\psec{Instruction}
Describe T1 and T2 as natural, continuous egocentric clips of about 15 seconds using visually observable information, along with optional conversation and mobile device notifications.
The intention should be inferable from T1, while T2 should provide an appropriate opportunity to act on it without directly revealing the intended action.
Any self-talk at T1 should describe only the current situation without explicitly stating the intention or future plan.
Avoid directly revealing the connection between T1 and T2.
\par

\psec{Input} \pvar{n}, \pvar{category}, \pvar{examples}\par

\psec{Output}
A structured JSON object containing the generated scenario pairs, including persona, delayed intention, T1 and T2 contexts, and the inference linking the two moments.\par
\end{promptbox}
\centering
\begin{promptbox}{Data Generation (Variants)}
\psec{Task}
Given a formation--decision pair, generate the additional assets needed to construct its controlled variants and long-horizon timelines without modifying the original pair.\par

\psec{Instruction}
Generate a negative decision context where the intention remains unresolved but cannot be acted on, an observable state-change event that resolves the intention before the original decision moment, and an ordinary intention-independent event to replace the formation moment.
Keep all scenes natural and grounded in observable context, while avoiding cues that reveal the intended action or duplicate the scripted events.\par

\psec{Input} \pvar{formation--decision pair}\par

\psec{Output}
A structured JSON object containing the generated variant-specific events.\par
\end{promptbox}
\caption{Summarized prompt for data generation.}
\label{fig:prompt_scenario}
\end{figure}

\noindent\textbf{Prompt for Data Generation.}
Fig.~\ref{fig:prompt_scenario} summarizes the prompts used for scenario generation in \benchmark.
To further diversify the generated data, we use multiple frontier LLMs, including GPT-4o~\cite{hurst2024gpt}, Claude-Sonnet-4.6~\cite{anthropic2026sonnet46}, and Gemini-2.5-Flash~\cite{comanici2025gemini}, and human annotators subsequently verify all generated samples.

\noindent\textbf{Prompt for LLM Judge.}
Fig.~\ref{fig:prompt_judge} shows the prompt used by the LLM judge for \texttt{Positive} instances.
The judge determines whether a response addresses the same intention as the reference and interprets relative time expressions based on the provided event timestamps.
Generic assistance that does not identify the delayed intention, such as ``You should take care of that'', is not considered sufficient.
For \texttt{State Change} instances, the judge additionally receives the state-change event and classifies the response as still pursuing the resolved intention, appropriately following up on the state change, or unrelated to either.

\begin{figure}[t]
\centering
\begin{promptbox}{LLM Judge}
\psec{Task}
Determine whether the assistant's response correctly conveys the reference delayed intention.\par

\psec{Instruction}
Judge whether the response refers to the same core action and specific matter as the reference intention, allowing paraphrases and omission of secondary details.
When the intention involves relative timing, resolve it using the timestamps and ensure that the response is temporally appropriate.
Generic reminders, status updates, related mentions without the intended action, or unspecific inquiries do not count as matches.\par

\psec{Input} \pvar{t1}, \pvar{t2}, \pvar{gold\_intention}, \pvar{assistance}\par

\psec{Output}
A JSON object containing a brief justification and whether the response matches the reference intention.\par
\end{promptbox}
\caption{Summarized judge prompt in the evaluation.}
\label{fig:prompt_judge}
\end{figure}

\section{Evaluation Metrics}
\label{sec:appendix_metrics}

We evaluate each method at the decision moment $e_d$ along three dimensions, including memory retrieval, proactive decision, and end-to-end assistance.
At $e_d$, the agent should provide proactive assistance for \texttt{Positive} timelines and remain silent for the other three variants.

\noindent\textbf{Memory Retrieval.}
Delayed intention recall (DIR) is computed on \texttt{Positive} instances and measures the fraction of gold delayed intentions whose formation moment $e_f$ is retrieved.
State-change recall (SCR) is computed on \texttt{State Change} instances and measures the fraction of timelines for which both the formation moment $e_f$ and corresponding state-change event $e_{sc}$ are retrieved for all intentions.
These two variants require recovering specific historical evidence for correct reasoning, while \texttt{Negative} and \texttt{No Formation} primarily evaluate whether the agent appropriately remains silent.
For RAG-based methods and \workname, retrieval is evaluated over the top-$k$ events passed to the reasoner.
For A-Mem and MemOS, an event is considered retrieved if a retrieved memory is constructed from it, while for T-Mem, if it is covered by a retrieved scene or item.
The retrieval metrics are not applicable to Mem0 because each memory update jointly represents multiple events, or to proactive agents and full-context LLMs, which do not explicitly retrieve past events.

\noindent\textbf{Response Judgment.}
For \texttt{Positive} instances, an LLM judge evaluates whether the response correctly addresses each gold delayed intention using the prompt in Fig.~\ref{fig:prompt_judge}.
A delayed intention counts as a true positive (TP) if the response provides appropriate assistance for the intention, and as a false negative (FN) otherwise, including when the agent remains silent.
For the other three variants, where the agent should remain silent at $e_d$, an unnecessary response counts as a false positive (FP), while correctly remaining silent is counted as a true negative (TN).
For \texttt{State Change} instances, if the agent responds, the judge further determines whether it appropriately follows up on the state-change event $e_{sc}$.
Such a follow-up is counted as a TN, while responses that still act on the resolved intention or are unrelated are counted as FPs.

\noindent\textbf{Proactive Decision.}
Following prior work on proactive agents~\cite{yang2025contextagent,lu2025proactive}, we evaluate proactive decision-making using the false detection rate (FD) and proactive prediction accuracy (Acc-P).
FD measures the rate of unnecessary assistance among instances where the agent should remain silent and is computed as
\begin{equation}
\mathrm{FD}=\frac{\mathrm{FP}}{\mathrm{FP}+\mathrm{TN}}.
\end{equation}
Acc-P measures whether the agent makes the correct proactive decision at $e_d$.
Both metrics evaluate the decision to provide assistance without considering the correctness of the assistance content.

\noindent\textbf{End-to-End Assistance.}
To further evaluate the content of proactive assistance, we report set-level precision, recall, and F1 based on TP, FN, and FP.
Since an instance may contain multiple gold delayed intentions at $e_d$, these metrics operate over the set of intentions, while for instances with a single intention, these metrics are the same as standard precision, recall, and F1.
They are computed as
\begin{equation}
\begin{gathered}
\mathrm{Set\text{-}P}=\frac{\mathrm{TP}}{\mathrm{TP}+\mathrm{FP}}, \quad
\mathrm{Set\text{-}R}=\frac{\mathrm{TP}}{\mathrm{TP}+\mathrm{FN}},\\
\mathrm{Set\text{-}F1}=
\frac{2\,\mathrm{Set\text{-}P}\,\mathrm{Set\text{-}R}}
{\mathrm{Set\text{-}P}+\mathrm{Set\text{-}R}}.
\end{gathered}
\end{equation}
Unlike Acc-P, proactive assistance accuracy (Acc-PA) further requires the provided assistance to be correct and is computed as
\begin{equation}
\mathrm{Acc\text{-}PA}
=\frac{\mathrm{TP}+\mathrm{TN}}
{\mathrm{TP}+\mathrm{FN}+\mathrm{FP}+\mathrm{TN}}.
\end{equation}
We use Set-F1 as the primary end-to-end metric because it jointly captures missed delayed intentions and unnecessary proactive responses.
All metrics are reported as percentages in the evaluation.

\section{Reliability of the LLM Judge}
\label{sec:appendix_judge}
\begin{table}[t]
\centering
\small
\setlength{\tabcolsep}{4pt}
\resizebox{\columnwidth}{!}{%
\begin{tabular}{lcccc}
\toprule
 & Annotator 1 & Annotator 2 & GPT-5-mini & GPT-5.6 Luna \\
\midrule
Annotator 1  & -    & 93.9 & 95.7 & 89.6 \\
Annotator 2  & 93.9 & -    & 94.8 & 95.7 \\
GPT-5-mini   & 95.7 & 94.8 & -    & 92.2 \\
GPT-5.6 Luna & 89.6 & 95.7 & 92.2 & -    \\
\bottomrule
\end{tabular}%
}
\caption{Pairwise agreement (\%) between two human annotators and two LLM judges.}
\label{tab:judge_agreement}
\end{table}

To validate the reliability of the LLM-as-a-judge evaluation for long-horizon proactive assistance, we evaluate judge robustness along two aspects, following prior work~\cite{li-etal-2026-locomo}, including alignment with human judgment and consistency across different judge LLMs.
We compare GPT-5-mini, the judge used in the experiments, against two independent human annotators and an additional frontier LLM judge, GPT-5.6 Luna.

\noindent\textbf{Response Sample.}
We sample 100 responses containing proactive assistance from results on \benchmark, with 20 responses from each method family, including proactive agents, full-context LLMs, RAG-based methods, memory systems, and \workname.

\noindent\textbf{Human Annotation.}
Two annotators who did not participate in benchmark construction independently evaluate all sampled responses.
They receive the same inputs as the LLM judge, without the LLM-as-a-judge outputs.

\noindent\textbf{Results.}
Table~\ref{tab:judge_agreement} shows strong agreement between the human annotators and the LLM judges.
The average agreement between GPT-5-mini and the two annotators is 95.3\%, indicating that the LLM judge used in our main evaluation closely aligns with human judgments. 
We further assess robustness to the choice of judge LLM using GPT-5.6 Luna.
Although its agreement with the human annotators is lower on average, it remains around 90\%.
Further analysis of the disagreements shows that GPT-5.6 Luna applies a stricter evaluation criterion, with most disagreements occurring when human annotators and GPT-5-mini accept a response but GPT-5.6 Luna marks it as unacceptable.
Overall, the high agreement and consistency between human annotators and LLMs indicate the reliability of LLM-based evaluation for our task.

\section{Full Results}
\label{sec:appendix_results}
\begin{table*}[t]
\centering
\small
\setlength{\tabcolsep}{4pt}
\resizebox{0.95\textwidth}{!}{%
\begin{tabular}{p{2.0cm}lcccccccc}
\toprule
& \textbf{Method}
& \multicolumn{2}{c}{\textbf{Memory Retrieval}}
& \multicolumn{2}{c}{\textbf{Proactive Decision}}
& \multicolumn{4}{c}{\textbf{End-to-End Assistance}} \\
\cmidrule(lr){3-4} \cmidrule(lr){5-6} \cmidrule(lr){7-10}
& & DIR$\uparrow$ & SCR$\uparrow$ & FD$\downarrow$ & Acc-P$\uparrow$ & Acc-PA$\uparrow$ & Set-P$\uparrow$ & Set-R$\uparrow$ & Set-F1$\uparrow$ \\
\midrule
\multirow{2}{=}{\centering Proactive Agents} & Proactive Agent~(8B) & - & - & 4.8 & 77.7 & 74.7 & 63.6 & 21.5 & 32.2 \\
 & ContextAgent~(8B) & - & - & 38.7 & 59.8 & 47.6 & 11.0 & 12.3 & 11.6 \\
\midrule
\multirow{10}{=}{\centering Full-Context\\LLMs\\(Open- and\\\mbox{Closed-Source})} & Qwen2.5-32B & - & - & \textbf{3.0} & 75.4 & 72.1 & 50.0 & 7.7 & 13.3 \\
 & Qwen3-4B & - & - & 19.6 & 67.9 & 62.2 & 23.3 & 15.4 & 18.5 \\
 & Qwen3-8B & - & - & 28.6 & 62.9 & 57.9 & 23.8 & 23.1 & 23.4 \\
 & Qwen3-14B & - & - & 12.5 & 74.1 & 69.1 & 40.0 & 21.5 & 28.0 \\
 & Qwen3-32B & - & - & 35.1 & 67.0 & 60.5 & 35.2 & 49.2 & 41.0 \\
 & Llama-3.3-70B & - & - & 7.1 & 74.6 & 71.2 & 45.5 & 15.4 & 23.0 \\
\cmidrule(l){2-10}
 & GPT-4o & - & - & 16.1 & 74.6 & 70.8 & 47.1 & 36.9 & 41.4 \\
 & GPT-4.1 & - & - & 9.5 & 79.0 & 75.5 & 60.0 & 36.9 & 45.7 \\
 & GPT-5.4-mini & - & - & \underline{4.2} & 75.4 & 72.5 & 53.3 & 12.3 & 20.0 \\
 & GPT-5.4 & - & - & 23.8 & 75.0 & 72.1 & 50.0 & \textbf{61.5} & 55.2 \\
\midrule
\multirow{3}{=}{\centering RAG-based Methods (GPT-5.4)} & all-MiniLM-L6-v2 & 10.8 & 8.9 & 4.8 & 74.1 & 71.2 & 42.9 & 9.2 & 15.2 \\
 & BM25 & 15.4 & 5.4 & 5.4 & 73.7 & 70.8 & 40.0 & 9.2 & 15.0 \\
 & text-embedding-3-small & 20.0 & 10.7 & 4.8 & 75.9 & 73.0 & 55.6 & 15.4 & 24.1 \\
\midrule
\multirow{5}{=}{\centering Memory Systems (GPT-5.4)} & Mem0 & - & - & 4.8 & 74.1 & 70.8 & 38.5 & 7.7 & 12.8 \\
 & A-Mem & 24.6 & \underline{19.6} & 8.3 & 75.4 & 72.5 & 51.7 & 23.1 & 31.9 \\
 & MemOS & 18.5 & 5.4 & 5.4 & 75.0 & 71.2 & 43.8 & 10.8 & 17.3 \\
 & T-Mem & \underline{30.8} & \underline{19.6} & 8.3 & 75.4 & 71.2 & 46.2 & 18.5 & 26.4 \\
 & T-Mem (Horizon-based) & 26.2 & 14.3 & 9.5 & 73.7 & 70.8 & 44.8 & 20.0 & 27.7 \\
\midrule
\multirow{2}{=}{\centering Ours} & \workname~(8B) & \textbf{83.1} & \textbf{66.1} & 7.7 & \underline{85.7} & \underline{82.8} & \underline{74.5} & \underline{58.5} & \underline{65.5} \\
 & \workname~(32B) & \textbf{83.1} & \textbf{66.1} & 5.4 & \textbf{88.4} & \textbf{85.4} & \textbf{81.6} & \textbf{61.5} & \textbf{70.2} \\
\bottomrule
\end{tabular}%
}
\caption{Main results on \benchmarklite. All metrics are in percentages (\%). The best and second-best results are in \textbf{bold} and \underline{underlined}, respectively. ``-'' denotes inapplicable retrieval metrics.}
\label{tab:appendix_lite}
\end{table*}

\noindent\textbf{Overall performance on \benchmarklite.}
Tab.~\ref{tab:appendix_lite} reports the complete results on \benchmarklite.
Compared with \benchmark, all methods show lower performance under raw sensory inputs.
This is because raw sensory observations are noisy, and context extraction may miss fine-grained details or introduce errors.
When accumulated over long histories, these imperfections further challenge long-context reasoning.
This degradation is particularly evident for several full-context LLMs, whose recall drops substantially under noisier sensory contexts.
In contrast, \workname maintains comparatively robust retrieval from context-aware memory, suggesting that prospective cues can remain effective when coarse contextual signals are preserved.
The remaining performance drop results from downstream proactive reasoning, where missing fine-grained information can make it harder to recognize and act on implicit delayed intentions.

\begin{table*}[t]
\centering
\small
\setlength{\tabcolsep}{4pt}
\resizebox{0.95\textwidth}{!}{%
\begin{tabular}{p{2.0cm}lcccccccc}
\toprule
& \textbf{Method}
& \multicolumn{2}{c}{\textbf{Memory Retrieval}}
& \multicolumn{2}{c}{\textbf{Proactive Decision}}
& \multicolumn{4}{c}{\textbf{End-to-End Assistance}} \\
\cmidrule(lr){3-4} \cmidrule(lr){5-6} \cmidrule(lr){7-10}
& & DIR$\uparrow$ & SCR$\uparrow$ & FD$\downarrow$ & Acc-P$\uparrow$ & Acc-PA$\uparrow$ & Set-P$\uparrow$ & Set-R$\uparrow$ & Set-F1$\uparrow$ \\
\midrule
\multirow{2}{=}{\centering Proactive Agents} & Proactive Agent~(8B) & - & - & 6.9 & 77.9 & 74.4 & 65.6 & 30.5 & 41.7 \\
 & ContextAgent~(8B) & - & - & 37.9 & 60.5 & 48.3 & 15.3 & 16.0 & 15.7 \\
\midrule
\multirow{10}{=}{\centering Full-Context\\LLMs\\(Open- and\\\mbox{Closed-Source})} & Qwen2.5-32B & - & - & 12.4 & 73.0 & 67.5 & 41.5 & 20.6 & 27.6 \\
 & Qwen3-4B & - & - & 13.4 & 70.8 & 64.8 & 30.5 & 13.7 & 18.9 \\
 & Qwen3-8B & - & - & 24.5 & 70.1 & 64.1 & 39.5 & 37.4 & 38.4 \\
 & Qwen3-14B & - & - & 13.4 & 74.3 & 68.9 & 46.8 & 27.5 & 34.6 \\
 & Qwen3-32B & - & - & 28.1 & 73.0 & 67.7 & 46.9 & 58.0 & 51.9 \\
 & Llama-3.3-70B & - & - & 18.6 & 69.6 & 64.8 & 37.4 & 26.0 & 30.6 \\
\cmidrule(l){2-10}
 & GPT-4o & - & - & 15.4 & 79.9 & 75.1 & 59.5 & 52.7 & 55.9 \\
 & GPT-4.1 & - & - & 10.5 & \underline{83.3} & \underline{79.2} & \underline{69.2} & 55.0 & 61.3 \\
 & GPT-5.4-mini & - & - & 6.2 & 76.0 & 70.7 & 53.7 & 16.8 & 25.6 \\
 & GPT-5.4 & - & - & 22.2 & 80.9 & 77.3 & 59.5 & \textbf{76.3} & \underline{66.9} \\
\midrule
\multirow{3}{=}{\centering RAG-based Methods (GPT-5.4)} & all-MiniLM-L6-v2 & 6.1 & 4.9 & \textbf{3.3} & 75.2 & 69.6 & 44.4 & 6.1 & 10.7 \\
 & BM25 & 13.7 & 7.8 & 4.9 & 75.0 & 70.3 & 51.6 & 12.2 & 19.8 \\
 & text-embedding-3-small & 13.7 & 8.8 & 5.9 & 75.2 & 69.6 & 47.1 & 12.2 & 19.4 \\
\midrule
\multirow{5}{=}{\centering Memory Systems (GPT-5.4)} & Mem0 & - & - & 4.9 & 75.2 & 68.9 & 40.0 & 7.6 & 12.8 \\
 & A-Mem & 22.9 & 17.6 & 6.5 & 77.0 & 71.6 & 57.4 & 20.6 & 30.3 \\
 & MemOS & 22.9 & 4.9 & 9.2 & 74.5 & 69.3 & 47.2 & 19.1 & 27.2 \\
 & T-Mem & 29.0 & 19.6 & 7.2 & 75.2 & 69.6 & 47.6 & 15.3 & 23.1 \\
 & T-Mem (Horizon-based) & \underline{55.7} & \underline{21.6} & 16.3 & 76.2 & 71.4 & 52.8 & 42.7 & 47.3 \\
\midrule
\multirow{2}{=}{\centering Ours} & \workname~(8B) & \textbf{80.2} & \textbf{69.6} & \underline{3.9} & \textbf{89.2} & \textbf{86.5} & \textbf{87.5} & \underline{64.1} & \textbf{74.0} \\
 & \quad w/o SFT & 80.2 & 69.6 & 17.6 & 77.9 & 72.8 & 55.0 & 50.4 & 52.6 \\
\bottomrule
\end{tabular}%
}
\caption{Out-of-domain results on \benchmark, with three categories excluded from reasoner fine-tuning.}
\label{tab:appendix_ood}
\end{table*}

\noindent\textbf{Out-of-Domain Evaluation.}
Table~\ref{tab:appendix_ood} shows the overall results of baseline methods and \workname~under the out-of-domain setting.
We split the eight everyday categories into five training categories and three categories excluded from reasoner fine-tuning.
Results indicate that \workname~maintains retrieval performance comparable to that under the main split, indicating that its memory retrieval remains effective across different scenario categories.
With the same retrieved context, fine-tuning improves Set-F1 by more than 20 points over the base Qwen3-8B reasoner on the held-out categories, showing that the learned proactive reasoning transfers beyond category-specific assistance patterns.
Notably, even without fine-tuning, \workname's retrieval combined with the base reasoner (w/o SFT) outperforms full-context Qwen3-8B by more than 10 points in Set-F1, further indicating the benefit of retrieving relevant historical context for long-horizon proactive assistance.
Overall, \workname retains its advantage over all baselines on unseen categories, consistent with the main evaluation.

\par\medskip\noindent\begin{minipage}{\columnwidth}
\centering
\small
\setlength{\tabcolsep}{3pt}
\resizebox{\columnwidth}{!}{%
\begin{tabular}{clccccccc}
\toprule
$m$ & \textbf{Backbone}
& \multicolumn{2}{c}{\textbf{\makecell{Memory\\Retrieval}}}
& \multicolumn{2}{c}{\textbf{\makecell{Proactive\\Decision}}}
& \multicolumn{3}{c}{\textbf{\makecell{End-to-End\\Assistance}}} \\
\cmidrule(lr){3-4} \cmidrule(lr){5-6} \cmidrule(lr){7-9}
& & DIR$\uparrow$ & SCR$\uparrow$ & FD$\downarrow$ & Acc-P$\uparrow$ & Set-P$\uparrow$ & Set-R$\uparrow$ & Set-F1$\uparrow$ \\
\midrule
4                  & GPT-4o-mini           & 81.3 & 71.0 & 4.3 & 89.0 & 85.7 & 63.4 & 72.9 \\
\multirow{2}{*}{8} & GPT-4o-mini & 81.3 & 72.0 & 4.3 & 90.0 & 86.3 & 66.7 & 75.2 \\
                   & GPT-4.1-mini          & 80.5 & 71.0 & 3.3 & 90.5 & 89.2 & 67.5 & 76.9 \\
12                 & GPT-4o-mini           & 81.3 & 73.0 & 3.0 & 90.5 & 89.9 & 65.0 & 75.5 \\
16                 & GPT-4o-mini           & 80.5 & 71.0 & 3.3 & 90.2 & 88.6 & 63.4 & 73.9 \\
\bottomrule
\end{tabular}%
}
\captionof{table}{Sensitivity of \workname's memory construction to the backbone LLM and the number of candidate facts $m$ considered for cross-event linking.}
\label{tab:analysis_linkm}
\end{minipage}\par\medskip

\noindent\textbf{Sensitivity to Memory Construction.}
Table~\ref{tab:analysis_linkm} shows \workname's results with different memory construction LLMs and candidate budgets $m$ for cross-event linking, using the same retrieval configuration and proactive reasoner.
When replacing the default memory construction model, GPT-4o-mini, with GPT-4.1-mini, \workname~achieves comparable retrieval and end-to-end performance, indicating robustness to the choice of construction model.
Performance also remains stable as $m$ varies from 4 to 16, showing that the design of cross-event linking is robust to the candidate budget.

\end{document}